\documentclass{article}

\usepackage{arxiv}

\usepackage[utf8]{inputenc} 
\usepackage[T1]{fontenc}    
\usepackage{hyperref}       
\usepackage{url}            
\usepackage{booktabs}       
\usepackage{amsfonts}       
\usepackage{nicefrac}       
\usepackage{microtype}      
\usepackage{xcolor}         

\usepackage[american]{babel}

\usepackage{amsmath}
\usepackage{amssymb}
\usepackage{amsthm}
\usepackage{tikz}
\usetikzlibrary{arrows,cd,decorations.markings,shapes.geometric,shapes}
\usepackage{mathtools}
\usepackage{ebproof}

\usetikzlibrary{arrows,cd,decorations.markings,shapes.geometric,shapes}

\usepackage{tikzit}

\tikzstyle{blackDot}=[inner sep=0mm, minimum size=1.5mm, draw=black, shape=circle, draw=white, fill=black, line width=0.2mm]
\tikzstyle{boxSquare}=[fill=white, draw=black, shape=rectangle, minimum width=7mm, minimum height=5mm, font={\scriptsize}]
\tikzstyle{boxBroad}=[fill=white, draw=black, shape=rectangle, minimum width=10mm, minimum height=5mm, font={\scriptsize}]
\tikzstyle{boxSmall}=[fill=white, draw=black, shape=rectangle, minimum width=5mm, minimum height=5mm, font={\scriptsize}]
\tikzstyle{emptyText}=[fill=none, draw=none, shape=circle, font={\tiny}]
\tikzstyle{emptyTextScripsize}=[fill=none, draw=none, shape=circle, font={\scriptsize}]
\tikzstyle{state}=[fill=white, draw=black, regular polygon, regular polygon sides=3, minimum width=0.8cm, shape border rotate=180, inner sep=0pt, font={\scriptsize}]
\tikzstyle{stateLarge}=[fill=white, draw=black, regular polygon, regular polygon sides=3, minimum width=1cm, shape border rotate=180, inner sep=0pt, font={\scriptsize}]
\tikzstyle{box very broad}=[fill=white, draw=black, shape=rectangle, minimum width=20mm]
\tikzstyle{stateReverse}=[fill=white, draw=black, regular polygon, regular polygon sides=3, minimum width=0.8cm, inner sep=0pt, font={\scriptsize}]
\tikzstyle{whiteDot}=[fill=white, draw=black, shape=circle, minimum size=1.2mm, shape=circle, inner sep=0mm]

\tikzstyle{edge}=[fill=none, draw=black, line width=0.65pt, -]
\tikzstyle{white edge}=[-, draw=white, line width=1.5mm]
\tikzstyle{ddashed}=[-, draw=black, dashed]
\tikzstyle{doubleline}=[-, double, draw=black, fill=none]
\tikzstyle{redLine}=[-, draw=red]
\tikzstyle{blueLine}=[-, draw=blue]

\usepackage[all,cmtip]{xy}

\usetikzlibrary{circuits.ee.IEC}

\tikzstyle{white dot}=[inner sep=0mm, minimum size=1.5mm, draw=black, shape=circle, text depth=-0.2mm, draw=black, fill=white, tikzit category=nodes]
\tikzstyle{black dot}=[inner sep=0mm, minimum size=1.5mm, draw=black, shape=circle, draw=black, fill=black, tikzit category=nodes]
\tikzstyle{observed}=[inner sep=0mm, minimum size=5mm, draw=black, shape=circle, text depth=-0.2mm, draw=white, tikzit draw=gray, fill=white, tikzit category=dag]
\tikzstyle{latent}=[inner sep=0mm, minimum size=5mm, draw=black, shape=circle, text depth=-0.2mm, draw=black, fill=white, tikzit category=dag]
\tikzstyle{small box}=[shape=rectangle, text height=1.5ex, text depth=0.25ex, yshift=0.5mm, fill=white, draw=black, minimum height=6mm, yshift=-0.5mm, minimum width=6mm, font={\small}, tikzit category=boxes]
\tikzstyle{medium box}=[shape=rectangle, draw=black, fill=white, small box, minimum width=8mm, tikzit category=boxes]
\tikzstyle{semilarge box}=[shape=rectangle, draw=black, fill=white, small box, minimum width=12.5mm, tikzit category=boxes]
\tikzstyle{large box}=[shape=rectangle, draw=black, fill=white, small box, minimum width=15mm, tikzit category=boxes]
\tikzstyle{upground}=[circuit ee IEC, thick, ground, rotate=90, scale=1.5, inner sep=-2mm, tikzit shape=circle, tikzit fill=blue, tikzit category=points]
\tikzstyle{downground}=[circuit ee IEC, thick, ground, rotate=-90, scale=1.5, inner sep=-2mm, tikzit shape=circle, tikzit fill=green, tikzit category=points]
\tikzstyle{point}=[regular polygon, regular polygon sides=3, draw, scale=0.75, inner sep=-0.5pt, minimum width=9mm, fill=white, regular polygon rotate=180, tikzit category=points]
\tikzstyle{copoint}=[regular polygon, regular polygon sides=3, draw, scale=0.75, inner sep=-0.5pt, minimum width=9mm, fill=white, tikzit category=points]
\tikzstyle{uniform}=[point, fill=gray, tikzit shape=circle, scale=0.5]
\tikzstyle{label}=[font={\footnotesize}, text height=1.5ex, text depth=0.25ex, tikzit draw=blue, tikzit fill=white, tikzit category=labels]
\tikzstyle{left label}=[label, anchor=east, xshift=2mm, tikzit draw=green, tikzit fill=white, tikzit category=labels]
\tikzstyle{right label}=[label, anchor=west, xshift=-2mm, tikzit draw=purple, tikzit fill=white, tikzit category=labels]
\tikzstyle{disintegration}=[draw=black, fill={gray!50}, tikzit fill=gray, shape=rectangle, minimum width=1.6cm, minimum height=1.2cm, opacity=0.3]
\tikzstyle{empty diag}=[shape=rectangle, draw=darkgray, dashed, minimum width=8mm, minimum height=8mm, yshift=0.5mm]

\tikzstyle{diredge}=[->, >=latex]
\tikzstyle{dashed edge}=[-, dashed]

\pgfdeclarelayer{edgelayer}
\pgfdeclarelayer{nodelayer}
\pgfsetlayers{edgelayer,nodelayer,main}
\tikzstyle{none}=[]

\tikzstyle{morphism}=[fill=white, draw=black, shape=rectangle]
\tikzstyle{medium box}=[fill=white, draw=black, shape=rectangle, minimum width=0.8cm, minimum height=0.9cm]
\tikzstyle{large morphism}=[fill=white, draw=black, shape=rectangle, minimum width=1.7cm, minimum height=1cm]
\tikzstyle{bn}=[fill=black, draw=black, shape=circle, inner sep=1.5pt]
\tikzstyle{state}=[fill=white, draw=black, regular polygon, regular polygon sides=3, minimum width=0.8cm, shape border rotate=180, inner sep=0pt]
\tikzstyle{medium state}=[fill=white, draw=black, regular polygon, regular polygon sides=3, minimum width=1.3cm, inner sep=0pt, shape border rotate=180]
\tikzstyle{large state}=[fill=white, draw=black, regular polygon, regular polygon sides=3, minimum width=2.2cm, shape border rotate=180, inner sep=0pt]
\tikzstyle{wn}=[fill=white, draw=black, shape=circle, inner sep=1.5pt]

\tikzstyle{arrow}=[->]
\tikzstyle{dashed box}=[-, dashed]

\tikzset{baseline=(current  bounding  box.center)}
\tikzset{every picture/.append style={scale=0.5}}

\usepackage{tikzit}

\usetikzlibrary{circuits.ee.IEC}

\usetikzlibrary{calc,arrows.meta}

\newcommand\cofib\rightarrowtail

\newcommand\mdel[1]{}

\usepackage{txfonts}
\usepackage{pxfonts}

\makeatletter
\newcommand{\xdashrightarrow}[2][]{\ext@arrow 0359\rightarrowfill@@{#1}{#2}}
\newcommand*{\doublerightarrow}[2]{\mathrel{
  \settowidth{\@tempdima}{$\scriptstyle#1$}
  \settowidth{\@tempdimb}{$\scriptstyle#2$}
  \ifdim\@tempdimb>\@tempdima \@tempdima=\@tempdimb\fi
  \mathop{\vcenter{
    \offinterlineskip\ialign{\hbox to\dimexpr\@tempdima+1em{##}\cr
    \rightarrowfill\cr\noalign{\kern.5ex}
    \rightarrowfill\cr}}}\limits^{\!#1}_{\!#2}}}
\newcommand*{\triplerightarrow}[1]{\mathrel{
  \settowidth{\@tempdima}{$\scriptstyle#1$}
  \mathop{\vcenter{
    \offinterlineskip\ialign{\hbox to\dimexpr\@tempdima+1em{##}\cr
    \rightarrowfill\cr\noalign{\kern.5ex}
    \rightarrowfill\cr\noalign{\kern.5ex}
    \rightarrowfill\cr}}}\limits^{\!#1}}}
\makeatother

\makeatletter
\newcommand{\twoarrows}[3][0.2ex]{%
  \mathrel{\mathpalette\twoarrows@{{#1}{#2}{#3}}}%
}
\newcommand{\twoarrows@}[2]{\twoarrows@@#1#2}
\newcommand{\twoarrows@@}[4]{%
  \vcenter{\offinterlineskip\m@th
    \ialign{\hfil##\hfil\cr
      $#1#3$\cr
      \noalign{\vskip#2}
      $#1#4$\cr
    }%
  }%
}
\makeatother

\newcommand{\beq}{\begin{equation}}
\newcommand{\eeq}{\end{equation}}

\newcommand{\cop}{\mathsf{copy}}
\newcommand{\del}{\mathsf{del}}

\newtheorem{theorem}{Theorem}
\newtheorem{definition}{Definition}

\newtheorem{example}{Example}

\usepackage[boxruled]{algorithm2e}
\usepackage{algorithmic}

\usepackage{multirow}
\usepackage{booktabs}
\usepackage{balance}
\usepackage{subfig}

\DeclareFontFamily{U}{dmjhira}{}
\DeclareFontShape{U}{dmjhira}{m}{n}{ <-> dmjhira }{}

\usepackage{natbib} 
\usepackage{mathtools} 
\usepackage{booktabs} 
\usepackage{tikz} 

\title{A Rose by Any Other Name Would Smell as Sweet: Categorical Homotopy Theory for Large Language Models\thanks{Draft under review.} }

\author{ Sridhar Mahadevan \\
	Adobe Research and University of Massachusetts, Amherst\\
	\texttt{smahadev@adobe.com, mahadeva@umass.edu}
}

\hypersetup{
pdftitle={A template for the arxiv style},
pdfsubject={q-bio.NC, q-bio.QM},
pdfauthor={David S.~Hippocampus, Elias D.~Striatum},
pdfkeywords={First keyword, Second keyword, More},
}

\begin{document}
\maketitle

\begin{abstract}
Natural language is replete with superficially different statements, such as ``Charles Darwin wrote" and ``Charles Darwin is the author of", which carry the same meaning. Large language models (LLMs) should generate the same next-token probabilities in such cases, but usually do not. Empirical workarounds have been explored, such as using $k$-NN estimates of sentence similarity to produce smoothed estimates.  In this paper, we tackle this problem more abstractly, introducing a categorical homotopy framework for LLMs. We introduce an LLM Markov category to represent probability distributions in language generated by an LLM, where the probability of a sentence, such as ``Charles Darwin wrote" is defined by an arrow in a Markov category. However, this approach runs into difficulties as language is full of equivalent rephrases, and each  generates a non-isomorphic  arrow in the LLM Markov category. To address this fundamental problem, we use categorical homotopy techniques to capture ``weak equivalences" in  an LLM Markov category.  We present a detailed overview of application of categorical homotopy to LLMs, from higher algebraic K-theory to model categories, building on powerful theoretical results developed over the past half a century. 
\end{abstract}


\newpage 

\tableofcontents

\newpage 

\section{Introduction}\label{sec:intro}

In recent years, large language models (LLMs) have revolutionized AI, building on the success of the Transformer model \citep{DBLP:conf/nips/VaswaniSPUJGKP17} and related structured state space sequence models \citep{DBLP:conf/iclr/GuGR22}.  LLMs have enabled building large foundation models using massive amounts of data \citep{fm}, which has implications for whether artificial general intelligence (AGI) \citep{agi-dm} may be achievable in the near term.  Recent advances, such as DeepSeek-R1 \citep{deepseekai2025deepseekr1incentivizingreasoningcapability} show that significant gains in the computational cost of building these models can be achieved by structuring the problem using trial-and-error search based on reinforcement learning \citep{DBLP:books/lib/SuttonB98}.  However, much remains unclear about exactly what these systems are learning. Also, existing theoretical results regarding LLMs, on the one hand show they cannot recognize Dyck languages or compute the parity function  \citep{DBLP:journals/tacl/Hahn20,merrill2022saturatedtransformersconstantdepththreshold}, and on the other hand show that they are a universal function approximator over sequences \citep{DBLP:conf/iclr/YunBRRK20} 

Our approach is inspired by categorical models of language \citep{asudeh,bradley:enriched-yoneda-llms,Coecke2019TheMO}, however these approaches do not address semantic homotopy: when do two seemingly different sentences in language  ``mean" the same thing? For example, \citet{bradley:enriched-yoneda-llms} define a syntactic category (see Definition~\ref{llm-syntax}) and a semantic category (see Definition~\ref{llm-semantics-defn}) for LLMs where objects are fragments of sentences $x$ and morphisms $x \rightarrow y$  represent possible extensions of fragment $x$ by $y$. The set of possible completions is represented as an object in an enriched monoidal preorder category of the unit interval $[0,1]$, where logical connectives such as implication are defined as an adjoint functors.  \citet{merrill2024learnsemanticsnextwordprediction} investigate whether implications are indeed learnable from co-occurrence patterns in LLMs. Neither of these studies focus on homotopy. A large body of work exists around modeling language with symmetric monoidal categories in {\em quantum NLP} \citep{coecke2020mathematics,coecke2010mathematicalfoundationscompositionaldistributional}, which converts natural language into circuits for  use with quantum computation \citep{Coecke_Kissinger_2017}.  A Python package exists called  {\tt lambeq} \citep{kartsaklis2021lambeqefficienthighlevelpython}.The quantum NLP work has not addressed the issue of homotopy. Their approach builds on a compositional theory of language defined by Lambek pregroup grammars, a type of category defined by a partially ordered monoid \citep{lambek}. 

LLMs encode sentence prefixes to a fixed-sized representations and  predict the next word in the text using this encoding. But, such next-token predictions are challenging for natural language paraphrases: for example, ``Charles Dickens is the author of" should generate the same next-token distribution as ``Charles Dickens wrote". \citet{khandelwal2020generalizationmemorizationnearestneighbor} use nearest-neighbor methods to improve an LLM's performance at this type of next-token prediction problem.  Our work proposes a rigorous homotopy theory to analyze similar text fragments. Our homotopic framework is also related to research on LLM model distillation \citep{hsieh2023distillingstepbystepoutperforminglarger}, but these approaches usually rely on supervised training to construct smaller task-specific models, whereas we are proposing an internal analysis of lower-dimensional manifolds that the model parameters lie on. Finally, our work is related to work that uses {\tt logprob} or {\tt logit} information from commercial LLM implementations to reveal internal architectural details \citep{finlayson2024logitsapiprotectedllmsleak} by finding linear lower-dimensional matrix manifolds that the logit scores lie on, but our approach uses more powerful nonlinear homotopic methods. 

To explain the main idea, we propose to model LLMs using the framework of a {\em Markov category} \citep{Fritz_2020}, a symmetric monoidal category with a comonoidal ``copy-delete" structure on each object. Such Markov categories have been shown recently to be capable of a wide range of intricate reasoning in probability and statistics, and a diagrammatic foundation that compares favorably in ease of use to the alternative ``assembly language" measure theory foundation \citep{halmos:book}. A probability distribution over a natural language phrase is encoded in a Markov category as the arrow 

\[ f: \psi \rightarrow \text{Charles Darwin wrote} \]

However, an immediate dilemma that an LLM Markov category by itself is incapable of recognizing that the paraphrase 

\[ g: \mu \rightarrow \text{Charles Darwin is the author of} \]

should be treated as ``isomorphic" to the previous arrow. In other words, in an LLM Markov category, many objects exist that appear syntactically different, but are semantically identical. We address this fundamental problem by introducing categorical homotopy techniques into LLMs, particularly the use of categorical techniques for modeling them. 

\section{Homotopy in LLMs}

\begin{figure}[t]
    \centering
    \includegraphics[width=0.4\linewidth]{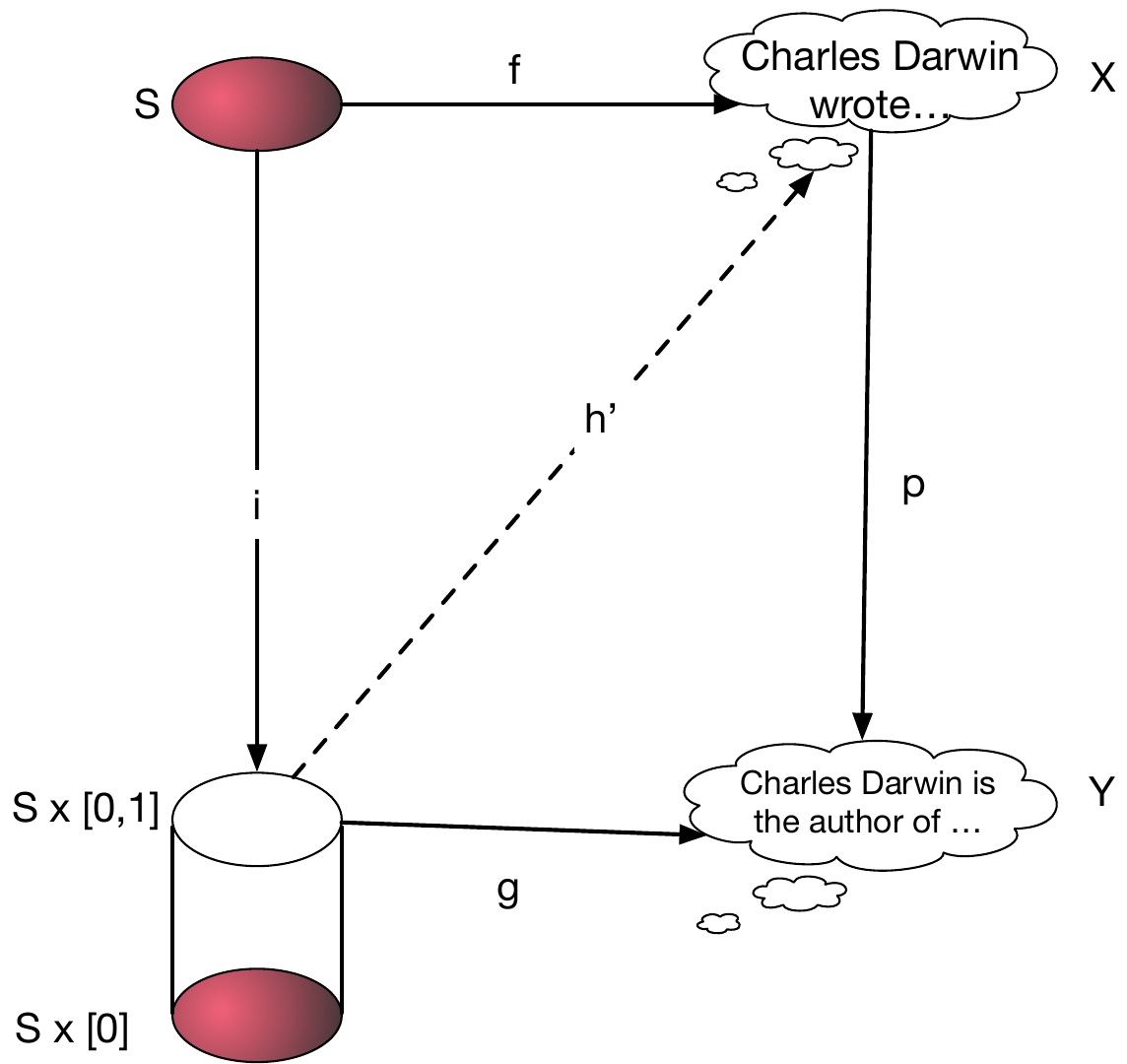}
    \caption{A lifting diagram for LLMs: any such commutative diagram has a ``lift" $h'$ if its composed morphisms $g \circ i = p \circ f$. Applied to LLMs, it asks whether the similarity between text sentences, i.e. is one fragment an equivalent rephrasing of another, can be ascribed topological ``meaning" in terms of how a circle fits into a cylinder when moved over the unit interval path $[0,1]$.}
    \label{fig:llm-lifting}
\end{figure}

Homotopy means deciding if two arbitrary objects are in some sense equivalent. For example, two real-valued functions $f,g: X \rightarrow \mathbb{R}$ are considered {\em homotopic}  if there is a continuous mapping $h: [0,1] \times X\rightarrow \mathbb{R}$  such that $h(0, x) = f(x)$ and $h(1,x) = g(x)$. If you imagine plotting $f$ and $g$, then the mental picture is of smoothly perturbing the graph of $f$ to align it with the graph of $g$. We want to define homotopy over LLM categories that capture ``syntax" and ``semantics" (Definition~\ref{llm-syntax} to Definition~\ref{llm-knn-cat}).  \citet{Quillen:1967} proposed {\em model categories} as a way to do abstract homotopy in any category, which we will review in Section~\ref{mc}. We prove in Section~\ref{sSets} that LLMs define model categories. 

The concept of {\em homotopy} originally arose in algebraic topology as a way to characterize when two spaces were essentially the same, such as a coffee cup is like a doughnut since they both have one hole, and they both are topologically distinct from a ``wedge sum of two circles" that looks like  a figure eight. Homotopy is usually defined in terms of {\em lifting diagrams} \citep{lifting} (see Figure~\ref{fig:llm-lifting}). \citet{SPIVAK_2013} formulates queries in relational databases using lifting diagrams. In that application, $X =$ {\tt I am flying...} is analogous to a database query in a language like SQL, $Y$ is the database itself from which we want answers (which presumably contains all possible completions of the query), and lifts represent potential answers to the query. 

\section{Category Theory for LLMs}

\label{introcat}

We introduce the framework of category theory \cite{maclane:71}, emphasizing its application to modeling LLMs. In simple terms, a category ${\cal C}$ is just a collection of {\em objects}  $c, d \in {\cal C}$ (of arbitrary complexity) and a collection of {\em arrows} ${\cal C}(c,d)$ for each pair of objects $c,d \in {\cal C}$. What makes something an object is up for grabs: in the case of LLMs, as we will see below, a natural way to define objects is as ``tokens" (roughly interpreted to mean words in a natural language). Naturally, words can be concatenated into sentences, which leads us to using {\em symmetric monoidal categories} \cite{fong2018seven}. Before we introduce these, we review some simple ways to model Transformer models as a category, and then review the work of \citep{bradley:enriched-yoneda-llms} on using enriched categories to model the next-token distribution for LLMs. In the next section, we formally introduce LLM Markov categories, which will be our primary framework in this paper. 

\begin{definition}
    A {\bf {category}} ${\cal C}$ is a graph ${\cal G}$ with two additional functions: {${\bf id}:$} $O \rightarrow A$, mapping each object $c \in C$ to an arrow {${\bf id}_c$} and $\circ: A \times_O A \rightarrow A$, mapping each pair of composable morphisms $\langle f, g \rangle$ to their composition $g \circ f$. 
\end{definition}

\subsection{Attention in Transformers} 
\label{attention} 

Central to the structure of Transformer is the computation of attention scores. A detailed review of some types of attention scores is given in \citep{chaudhari2021attentivesurveyattentionmodels}. Ignoring many details, the overall idea is to compute some type of ``dot product" between two tokens that are embedded in Euclidean space. We provide a brief review of some approaches, and then define a category of Transformer models. Especially noteworthy is the fact that Transformers intrinsically compute permutation-equivariant functions \citep{DBLP:conf/iclr/YunBRRK20}. To extend this basic symmetric computation of attention scores to nonsymmetric natural language applications, the inputs are weighted in some fashion, using a sinusoidal function in the original Transformer model \citep{DBLP:conf/nips/VaswaniSPUJGKP17}, and later studied in a more general Relative Positional Encoding framework \citep{shaw2018selfattentionrelativepositionrepresentations}. 
The general structure of an attention model is given as 
\[ A(q, K, v) = \sum_i p(a(k_i, q)) \times v_i \]
where $K$ and $V$ are key-value pairs, $q$ is a query, $a$ is an alignment function, and $p$ is a distribution function. Some sample alignment functions are given in Table~\ref{tab:alignmentfn}.

\begin{table}[!h]
\centering
\begin{tabular}{ll}
\hline
Function  & Equation \\
\hline \hline
similarity & \(\displaystyle a(k_i, q) = sim(k_i, q) \) \\
dot product & \(\displaystyle a(k_i, q) = q^{T}k_i \)  \\
scaled dot product & \(\displaystyle a(k_i, q) = \frac{q^{T}k_i}{\sqrt{d_{k}}} \) \\
general & \(\displaystyle a(k_i, q) = q^{T}Wk_i \) \\
biased general & \(\displaystyle a(k_i, q) = k_i(Wq + b) \)  \\
activated general & \(\displaystyle a(k_i, q) = act(q^{T}Wk_i + b) \)  \\
generalized kernel & \(\displaystyle a(k_i, q) = \phi(q)^{T}\phi(k_i) \) \\
\hline \hline
\end{tabular}
\begin{tabular}{p{14 cm}}
\end{tabular}
\vspace*{0.5 cm}
\caption{Some Alignment Functions \citep{chaudhari2021attentivesurveyattentionmodels}.}
\label{tab:alignmentfn}
\end{table}

\subsection{Transformers as a Category} 
\label{transcat}

To help concretize the more abstract presentation in the main paper, let us define a category of Transformer models. It is straightforward to extend our categorical model  below to  other building blocks of generative AI systems, including structured state space sequence models \citep{DBLP:conf/iclr/GuGR22} or image diffusion models \citep{DBLP:conf/nips/SongE19}. As with all generative AI systems, the fundamental structure of a Transformer model is a compositional structure made up of modular components, each of which computes a {\em permutation-equivariant} function over the vector space $\mathbb{R}^{d \times n}$ of $n$-length sequences of tokens, each embedded in a space of dimension $d$. We can define a commutative diagram showing the permutation equivariant property as shown below. 

To begin with, following \citep{DBLP:conf/lics/FongST19}, we can generically  define a neural network layer of type $(n_1, n_2)$ as a subset $C \subseteq [n_1] \times [n_2]$ where $n_1, n_2 \in \mathbb{N}$ are natural numbers, and $[n] = \{1, \ldots, n \}$. These numbers $n_1$ and $n_2$ serve to define the number of inputs and outputs of each layer, $C$ is a set of connections, and $(i, j) \in C$ means that node $i$ is connected to node $j$ in the network diagram. It is straightforward to define activation functions $\sigma: \mathbb{R} \rightarrow \mathbb{R}$ for each layer, but essentially each network layer defines a parameterized function $I: \mathbb{R}^{|C| + n_2} \times \mathbb{R}^{n_1} \rightarrow \mathbb{R}^{n_2}$, where the $\mathbb{R}^|C|$ define the edge weights of each network edge and the $\mathbb{R}^{n_2}$ factor encodes individual unit biases. 

We can specialize these to Transformer models, in particular, noting that the Transformer models compute specialized types of permutation-equivariant functions as defined by the commutative diagram below (in practice, the Transformer model adds an external positional encoding, such as the one defined in \citep{DBLP:conf/nips/VaswaniSPUJGKP17} or later variants, such as Relative Positional Embedding \citep{shaw2018selfattentionrelativepositionrepresentations}, which we will discuss below). 

\[\begin{tikzcd}
	X && Y && Z \\
	\\
	XP && YP && ZP
	\arrow["f", from=1-1, to=1-3]
	\arrow["P", from=1-3, to=3-3]
	\arrow["P"', from=1-1, to=3-1]
	\arrow["f"', from=3-1, to=3-3]
	\arrow["g", from=1-3, to=1-5]
	\arrow["g"', from=3-3, to=3-5]
	\arrow["P", from=1-5, to=3-5]
\end{tikzcd}\]

In the above commutative diagram, vertices are objects, and arrows are morphisms that define the action of a Transformer block. Here, $X \in \mathbb{R}^{d \times n}$ is a $n$-length sequence of tokens of dimensionality $d$. $P$ is a permutation matrix. The function $f$ computed by a Transformer block is such that $f(XP) = f(X) P$. This property is defined in the above diagram by setting $Y = f(X) P$, which can be computed in two ways, either first by permuting the input by the matrix $P$, and then applying $f$, or by 

Let us understand the permutation equivariant property of the Transformer model in a bit more detail. Our notation for the Transformer model is based on \citep{DBLP:conf/iclr/YunBRRK20}, although there are many variations in the literature that we do not discuss further. These can be adapted into our categorical framework fairly straightforwardly based on the approach outlined below. Transformer models are inherently compositional, which makes them particularly convenient to model using category theory. 
\begin{definition}
    A {\bf Transformer} block is a sequence-to-sequence function mapping $\mathbb{R}^{d \times n} \rightarrow \mathbb{R}^{d \times n}$. There are generally two layers: a {\em self-attention} layer and a token-wise feedforward layer. We assume tokens are embedded in a space of dimension $d$. Specifically, we model the inputs $X \in \mathbb{R}^{d \times n}$ to a Transformer block as $n$-length sequences of tokens in $d$ dimensions, where each block computes the following function defined as $t^{h,m,r}: \mathbb{R}^{d \times n}: \mathbb{R}^{d \times n}$: 

\begin{eqnarray*}
    \mbox{Attn}(X) &=& X + \sum_{i=1}^h W^i_O W^i_V X \cdot \sigma[W^i_K X)^T W^i_Q X] \\
    \mbox{FF}(X) &=& \mbox{Attn}(X) + W_2 \cdot \mbox{ReLU}(W_1 \cdot \mbox{Attn}(X) + b_1 {\bf 1}^T_n, 
\end{eqnarray*}

where $W^i_O \in \mathbb{R}^{d \times n}$, $W^i_K, W^i_Q, W^i_Q \in \mathbb{R}^{d \times n}$, $W_2 \in \mathbb{R}^{d \times r}$, $W_1 \in \mathbb{R}^{r \times d}$, and $b_1 \in \mathbb{R}^r$. The output of a Transformer block is $FF(X)$. Following convention, the number of ``heads" is $h$, and each ``head"  size $m$
are the principal parameters of the attention layer, and the size of the ``hidden" feed-forward layer is $r$. 
\end{definition}
Transformer models take as input objects $X \in \mathbb{R}^{d \times n}$ representing $n$-length sequences of tokens in $d$ dimensions, and act as morphisms that represent permutation equivariant functions $f: \mathbb{R}^{d \times n} \rightarrow \mathbb{R}^{d \times n}$ such that $f(XP) = f(X)P$ for any permutation matrix.  \citet{DBLP:conf/iclr/YunBRRK20} show that the actual function computed by the Transformer model defined above is a permutation equivariant mapping. 

Concretely, we define a category of transformers ${\cal C}_{T}$  where the objects are vectors $x \in \mathbb{R}^{d \times n}$ representing sequences of $d$-dimensional tokens of length $n$, and the composable arrows are {\em permutation-equivariant} functions ${\cal T}^{h,m,r}$ comprised of a composition of  transformer blocks $t^{h,m,r}$ of $h$ heads of size $m$ each, and a feedforward layer of $r$ hidden nodes.  Objects in a category interact with each other through arrows or morphisms. In the category ${\cal C}_T$ of Transformer models, the morphisms are the equivariant maps $f$ by which one Transformer model block can be composed with another. 

\begin{definition}
    The category ${\cal C}_T$ of Transformer models: 
    \begin{itemize}
        \item The objects $\mbox{Obj({\cal C})}$ are defined as vectors $X \in \mathbb{R}^{d \times n}$ denoting $n$-length sequences of tokens of dimension $d$. 

        \item The arrows or morphisms of the category ${\cal C}_T$ are defined as a family of sequence-to-sequence functions and defined as: 

        \[ T^{h,m,r} \coloneqq \{f: \mathbb{R}^{d \times n} \rightarrow \mathbb{R}^{d \times n} \ | f(XP) = f(X) P \} \]
    \end{itemize}
    where $P$ is some permutation matrix, $h$ is the number of heads of size $m$ each, and each feedforward layer has $r$ hidden nodes. 
\end{definition}
To overcome the restriction to permutation-equivariant functions, it is common in Transformer implementations to include a Relative Position Embedding function (RPE) \citep{shaw2018selfattentionrelativepositionrepresentations}.

 \subsection{LLM Next-token Distributions as Categories}
\label{llmcat-defns}
Now, we define categories for LLMs based on next-token distributions \citep{bradley:enriched-yoneda-llms}, and introduce a hybrid category based on the $k$-NN LLM model proposed by \citet{khandelwal2020generalizationmemorizationnearestneighbor}.
\begin{definition}\citep{bradley:enriched-yoneda-llms}
\label{llm-syntax}
    The {\bf LLM syntax category}  ${\cal L}$ of a large language model is defined as a category enriched over a monoidal preorder $[0,1]$, whose objects are strings in a natural language, and whose morphisms are defined as ${\cal L}(x, y) \coloneqq P(y | x) $,   which means that if $y$ is an expression such as ``I am flying to San Diego", which extends the expression ``I am flying", then $P(y | x)$ gives the conditional probability of such an extension (modulo a training set over which the model was trained). 
\end{definition}
To define more ``semantic" categories for LLMs, we use {\em enriched copresheaves}: these are Yoneda embeddings where a token $x$ is mapped into the set of all possible completions ${\cal L}(x,-)$ -- the copresheaf -- enriched over  the unit interval $[0,1]$. The Yoneda embedding maps an object $x$ to a covariant set-valued ${\cal L}(x, -)$ in general, but in special cases, such as for LLMs, this functor is itself an object of an enriched category, such as a symmetric monoidal preorder over the unit interval $[0,1]$.  We can consider the Yoneda embedding of the LLM category ${\cal L}$ to be a ``semantic" category that defines the meaning of sentences. 
\begin{definition}\citep{bradley:enriched-yoneda-llms}
\label{llm-semantics-defn}
For the LLM category ${\cal L}$ in Definition~\ref{llm-syntax}, the {\bf semantic category} $\hat{{\cal L}} \coloneqq [0, 1]^{\cal L}$ is the $[0,1]$-enriched category of $[0,1]$-enriched copresheaves on the $[0,1]$-category ${\cal L}$. 
\end{definition}
Finally, let us define a new type of category based on the $k$-NN large language model defined by \citet{khandelwal2020generalizationmemorizationnearestneighbor} (see Section~\ref{knn-trans} for additional details). 
\begin{definition}
\label{llm-knn-cat}
    The {\bf $k$-NN LLM  syntax category}  ${\cal L}_{kNN}$ is defined as a category whose morphisms ${\cal L}_{kNN}(y | x)$ are based on combining a synthetic $k$-NN computed probability from a datastore ${\cal D}$ and the LLM learned next-token distribution, defined as: 
 \[ p(y|x) = \lambda p_{KNN}(y | x) + (1 - \lambda) p_{LM}(y | x) \]
 computed by querying  a datastore $\langle {\cal K}, {\cal V} \rangle$ with an internal self-attention function computed by the Transformer model $f(x)$ to retrieve its $k$ nearest neighbors ${\cal N}$ using some distance function $d(., .)$, and using a softmax of their negative distances: 
 \[ p_{KNN}(y | x) \propto \sum_{\langle k_i, v_i \rangle \in {\cal N}} {\bf 1}_{y = v_i} \exp{(-d(k_i, f(x)))}\]
The {\bf $k$NN LLM semantics category} is then $\hat{{\cal L}}_{kNN} \coloneqq [0, 1]^{{\cal L}_{kNN}}$ is the $[0,1]$-enriched category of $[0,1]$-enriched copresheaves on the $[0,1]$-category ${\cal L}_{kNN}$. 
\end{definition}
\citet{khandelwal2020generalizationmemorizationnearestneighbor} shows that $k$-NN LLMs have a higher perplexity score than regular LLMs precisely because they  model more accurately the next-token distribution over equivalent phrases, such as {\tt Charles Dickens wrote} and {\tt Charles Dickens is the author of}.

\subsection{Synthetic Token Probabilities using Nearest-Neighbor Language Modeling }
\label{knn-trans}

\cite{khandelwal2020generalizationmemorizationnearestneighbor} proposed $k$-NN nearest neighbor language models, which combine a {\em synthetic probability} constructed from key-query value pairs $\langle k_i v_i \rangle$ from each training example $\langle c_i, w_i \rangle$, where $c_i$ is a {\em context-phrase}, such as {\tt Charles Dickens wrote}, and $f(.)$ is a function that maps a context $c$ to a fixed-length vector representation computed by a pre-trained Transformer model. For example, $f(c)$ could map $c$ to a fixed-length vector output by some self-attention layer (which we described in the previous section). The datastore $\langle {\cal K}, {\cal V} \rangle$ is the set of all key-value pairs constructed from all the training examples in ${\cal D}$: 
\[ \langle {\cal K}, {\cal V} \rangle = \{(f(c_i), w_i) | (c_i, w_i) \in {\cal D} \} \]
 At the time of testing, the language model learned by an LLM generates the output distribution $p_{LM}(y | x)$, and queries the datastore $\langle {\cal K}, {\cal V} \rangle$ with $f(x)$ to retrieve its $k$ nearest neighbors ${\cal N}$ using some distance function $d(., .)$. A synthetic probability is then computed over the neighbors using a softmax of their negative distances: 
 \[ p_{KNN}(y | x) \propto \sum_{\langle k_i, v_i \rangle \in {\cal N}} {\bf 1}_{y = v_i} \exp{(-d(k_i, f(x)))}\]
 The final distribution for the $k$-NN LLM is then given by 
 \[ p(y|x) = \lambda p_{KNN}(y | x) + (1 - \lambda) p_{LM}(y | x) \]
 Thus, we can define a new LLM category for $k$-NN LLMs in this way called ${\cal L}_{knn}$ where the next-token probabilities are a linear interpolation of the learned probabilities combined with a nearest-neighbor prediction. \citet{khandelwal2020generalizationmemorizationnearestneighbor} show a range of empirical results on large datasets showing that this interpolated probability improves the prediction capability of LLMs with a higher perplexity score. 

\subsection{The Calculus of Ends and Coends} 
\label{coends}

We can define the space of natural transformations between two functors $F, G: {\cal L}_s \Rightarrow {\cal L_m}$  that map between the ``syntax" category of LLMs and its ``semantic" category in terms of the calculus of (co) ends \citep{loregian_2021}.  The category of {\em wedges} are defined by a collection of objects comprised of bifunctors $F: {\cal C}^{op} \times C \rightarrow {\cal D}$, and a collection of arrows between each pair of bifunctors $F, G$ called a {\em dinatural transformation} (as an abbreviation for diagonal natural transformation). We will see below that the initial and terminal objects in the category of wedges correspond to a beautiful idea first articulated by Yoneda called the {\em coend} or {\em end} \citep{yoneda-end}. \citet{loregian_2021} has an excellent treatment of coend calculus, which we will use below. 

\begin{definition}
    Given a pair of bifunctors $F, G: {\cal C}^{op} \times {\cal C} \rightarrow {\cal D}$, a {\bf dinatural transformation} is defined as follows: 

\[\begin{tikzcd}
	&& {F(c',c)} \\
	{F(c,c)} &&&& {F(c',c')} \\
	\\
	{G(c,c)} &&&& {G(c',c')} \\
	&& {G(c,c')}
	\arrow["{F(f,c)}", from=1-3, to=2-1]
	\arrow["{F(c',f)}"', from=1-3, to=2-5]
	\arrow[dashed, from=2-1, to=4-1]
	\arrow[dashed, from=2-5, to=4-5]
	\arrow["{G(c,f)}", from=4-1, to=5-3]
	\arrow["{G(f,c)}"', from=4-5, to=5-3]
\end{tikzcd}\]

\end{definition}

As \citep{loregian_2021} observes, just as a natural transformation interpolates between two regular functors $F$ and $G$ by filling in the gap between their action on a morphism $Ff$ and $Fg$ on the codomain category, a dinatural transformation ``fills in the gap" between the top of the hexagon above and the bottom of the hexagon. 

We can define a {\em constant bifunctor} $\Delta_d: {\cal C}^{op} \times {\cal C} \rightarrow {\cal D}$ by the object it maps everything to, namely the input pair of objects $(c, c') \rightarrow d$ are both mapped to the object $d \in {\cal D}$, and the two input morphisms $(f, f') \rightarrow {\bf 1}_d$ are both mapped to the identity morphism on $d$. We can now define {\em wedges} and {\em cowedges}. 

\begin{definition}
    A {\bf wedge} for a bifunctor $F: {\cal C}^{op} \times {\cal C} \Rightarrow {\cal D}$ is a dinatural transformation $\Delta_d \rightarrow F$ from the constant functor on the object $d \in {\cal D}$ to $F$. Dually, we can define a {\bf cowedge} for a bifunctor $F$ by the dinatural transformation $P \Rightarrow \Delta_d$. 
\end{definition}

We can now define a {\em category of wedges}, each of whose objects are wedges, and for arrows, we choose arrows in the co-domain category that makes the diagram below commute. 

\begin{definition}
    Given a fixed bifunctor $F: {\cal C}^{op} \times {\cal C} \rightarrow {\cal D}$, we define the {\bf category of wedges} ${\cal W}(F)$ where each object is a wedge $\Delta_d \Rightarrow F$ and given a pair of wedges $\Delta_d \Rightarrow F$ and $\Delta_d' \Rightarrow F$, we choose an arrow $f: d \rightarrow d'$ that makes the following diagram commute: 

\[\begin{tikzcd}
	d &&&& {d'} \\
	\\
	&& {F(c,c)}
	\arrow["f", from=1-1, to=1-5]
	\arrow["{\alpha_{cc}}"', from=1-1, to=3-3]
	\arrow["{\alpha'_{cc}}", from=1-5, to=3-3]
\end{tikzcd}\]
Analogously, we can define a {\bf category of cowedges} where each object is defined as a cowedge $F \Rightarrow \Delta_d$. 
\end{definition}
With these definitions in place, we can once again define the universal property in terms of initial and terminal objects.
\begin{definition}
    Given a bifunctor $F: {\cal C}^{op} \times {\cal C} \rightarrow {\cal D}$, the {\bf end} of $F$ consists of a terminal wedge $\omega: \underline{{\bf end}}(F) \Rightarrow F$. The object $\underline{{\bf end}}(F) \in D$ is itself called the end. Dually, the {\bf coend} of $F$ is the initial object in the category of cowedges $F \Rightarrow \underline{{\bf coend}}(F)$, where the object $\underline{{\bf coend}}(F) \in {\cal D}$ is itself called the coend of $F$.  
\end{definition}

Remarkably, probabilities can be formally shown to define ends of a category \citep{Avery_2016}, and topological embeddings of datasets, as implemented in popular dimensionality reduction methods like UMAP \citep{umap}, correspond to coends \citep{maclane:71}.  These connections suggest the canonical importance of the category of wedges and cowedges  in developing a comprehensive understanding of LLMs.

 \subsection{Enriched LLM Categories}

Let us define a few properties of enriched categories for LLMs next. Usually, we are interested in the case when ${\cal V}$ is closed, and thus enriched over itself, in which case the above relationship simplifies to ${\cal C}(x, y) \leq {\cal V}(fx, fy)$. 

 \begin{definition}
     If ${\cal C}$ is a category enriched over a commutative monoidal preorder ${\cal V}$, then an {\bf enriched copresheaf} is a function $f: {\cal C} \rightarrow {\cal V}$ satisfying ${\cal C}(x,y) \leq {\cal V}(fx, fy)$ over all objects $x$ and $y$ in ${\cal C}$.  For the category of LLMs, $y$ will represent an extension of the text fragment $x$, and ${\cal V}(fx, fy)$ defines the probability of extension. \end{definition}

We can actually define a topos \citep{maclane1992sheaves} over the enriched LLM category, since it possesses  internal exponential objects, as every topos contains such objects. We can use the following definition (where the notion of an {\em enriched end} is defined at greater length in Section~\ref{coends} -- see also \citep{loregian_2021}). 
\begin{definition}
    The ${\cal V}$-enriched functor category ${\cal D}^{\cal C}$, whose objects are ${\cal V}$ functors from ${\cal C}$ to ${\cal D}$, has an internal hom object ${\cal D}^{\cal C}(f,g)$ between any two functors $f,g \in {\cal D}^{\cal C}$ is defined as the following enriched end \citep{loregian_2021}, which always exists since we assume ${\cal V}$ has all limits. 
\[ {\cal D}^{\cal C}(f, g) = \int_{c \in {\cal C}} {\cal D}(fc, gc) \]
\end{definition}
For the special case that ${\cal V}$ is the symmetric monoidal preorder over the unit interval, this enriched coend turns into something much simpler. 
\begin{theorem}\citep{bradley:enriched-yoneda-llms}
If ${\cal C}$ is a category enriched over $[0,1]$, then the category $\hat{C} = [0,1]^{\cal C}$ of copresheaves is also enriched over $[0,1]$. The $[0,1]$ object between any pair of copresheaves $f, g: {\cal C} \rightarrow [0,1]$ is given as the following minimization over all objects in ${\cal C}$. 
\[ \hat{C}(f, g) = \inf_{c \in {\cal C}} \{ [fc, gc] \} \]
\end{theorem}
The internal hom object $[x,y] \in {\cal V}$ is defined as truncated division, the above expression significantly simplifies to the following: 
\[ \hat{C}(f, g) = \inf_{c \in {\cal C}} \{ 1, \frac{gc}{fc} \} \]
\begin{theorem}\citep{bradley:enriched-yoneda-llms}
If ${\cal C}$ is a category enriched over the symmetric monoidal preorder ${\cal V} = [0,1]$, then the Yoneda embedding ${\cal C}(x, -) $ is also a $[0,1]$-enriched functor. 
\end{theorem}
These definitions and results essentially can be viewed as a special case of a metric Yoneda Lemma in generalized metric spaces, where the unit interval $[0,1]$ is mapped to non-negative distances in $[0,\infty]$ using the $-\log$ mapping. 
\begin{theorem}\citep{bradley:enriched-yoneda-llms}
    For any object $x$ in a $[0,1]$-enriched category ${\cal C}$, and any $[0,1]$-copresheaf $f: {\cal C} \rightarrow [0,1]$, the Yoneda embedding once again serves as the representer of evaluation in the unit interval space, namely $\hat{C}({\cal C}(x, -), f) = f(x)$. 
\end{theorem}
\begin{theorem}\citep{bradley:enriched-yoneda-llms}
For all objects $x$ and $y$ in a $[0,1]$-enriched category ${\cal C}$, there is the following isomorphism: 
  \[   {\cal C}(y,x) = \hat{C}({\cal C}(x, -), {\cal C}(y, -)) \]
\end{theorem}

\begin{definition}\citep{bradley:enriched-yoneda-llms}
For the LLM category ${\cal L}$ in Definition~\ref{llm-syntax}, the {\bf semantic category} $\hat{{\cal L}} \coloneqq [0, 1]^{\cal L}$ is the $[0,1]$-enriched category of $[0,1]$-enriched copresheaves on the $[0,1]$-category ${\cal L}$. 
\end{definition}
\begin{definition}
    An {\bf enriched functor}  $f: {\cal C} \rightarrow {\cal D}$ for categories enriched over a symmetric monoidal preorder $({\cal V}, \otimes, \leq)$ is a function ${\cal C} \rightarrow {\cal D}$ that satisfies the relation ${\cal C}(x, y) \leq {\cal D}(fx, fy)$. 
\end{definition}

\section{Monoidal LLM Categories} 

In this section, we present a categorical framework for analyzing LLMs based on a symmetric monoidal category representation called a {\em Markov category} \citep{Fritz_2020}. We first give a brief explanation of symmetric monoidal categories, before more formally introducing Markov categories. We compare our approach to previous work on categorical representations of LLMs \citep{bradley:enriched-yoneda-llms}. 

\subsection{Symmetric Monoidal Categories}
\label{smc}

\begin{figure}
    \centering
    \includegraphics[width=0.75\linewidth]{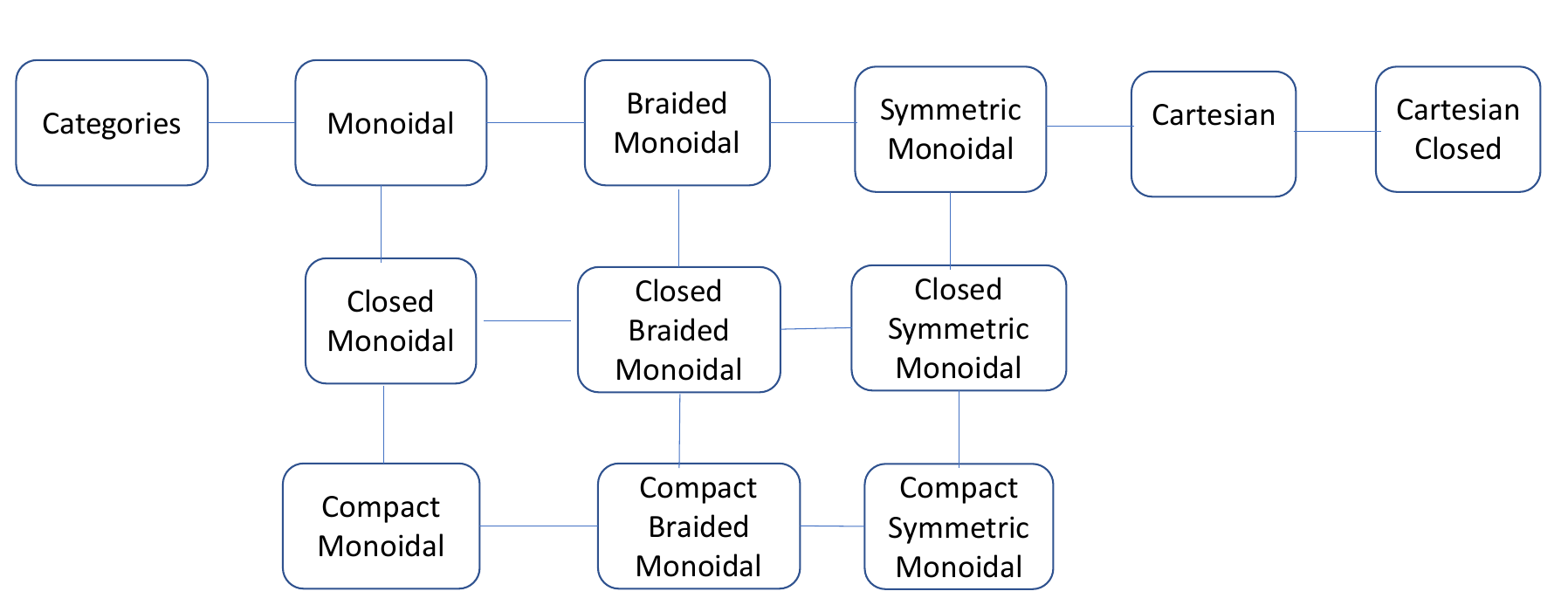}
    \caption{Structure of monoidal categories.}
    \label{monoidal-cats}
\end{figure}

Categorical models of causality \citep{fong:ms,fritz:jmlr,string-diagram-surgery,DBLP:journals/entropy/Mahadevan23} are usually defined over symmetric monoidal categories, which we briefly review now \citep{maclane:71,richter2020categories} (see Figure~\ref{monoidal-cats}). 

\begin{definition}
    A {\bf monoidal category} is a category {\cal C} together with a functor $\otimes: {\cal C} \times {\cal C} \rightarrow {\cal C}$, an identity object $e$ of {\cal C} and natural isomorphisms $\alpha, \lambda, \rho$ defined as: 

    \begin{eqnarray*}
        \alpha_{C_1, C_2, C_3}: C_1 \otimes (C_2 \otimes C_3) & \cong & (C_1 \otimes C_2) \otimes C_2,\\
        \lambda_C: e \otimes C & \cong & C,   \\
        \rho: C \otimes e & \cong & C, 
    \end{eqnarray*}
\end{definition}

The natural isomorphisms must satisfy coherence conditions called the ``pentagon" and ``triangle" diagrams \citep{maclane:71}. An important result shown in \citep{maclane:71} is that these coherence conditions guarantee that all well-formed diagrams must commute.  There are many natural examples of monoidal categories, the simplest one being the category of finite sets, termed {\bf FinSet} in \citep{Fritz_2020}, where each object $C$ is a set, and the tensor product $\otimes$ is the Cartesian product of sets, with functions acting as arrows. Deterministic causal models can be formulated in the category {\bf FinSet}. Other examples include the category of sets with relations as morphisms, and the category of Hilbert spaces \citep{Heunen2019}. Markov categories \citep{fritz:jmlr} are monoidal categories, where the identity element $e$ is also a terminal object, meaning there is a unique ``delete" morphism $d_e: X \rightarrow e$ associated with each object $X$. \citep{Fritz_2020} shows they form a unifying foundation for probabilistic and statistical reasoning. 
\begin{definition}
    A {\bf symmetric monoidal category} is a monoidal category $({\cal C}, \otimes, e, \alpha, \lambda, \rho)$ together with a natural isomorphism 
    \begin{eqnarray*}
       \tau_{C_1, C_2}: C_1 \otimes C_2 \cong C_2 \otimes C_1, \ \ \mbox{for all objects} \ \ C_1, C_2
    \end{eqnarray*}
    where $\tau$ satisfies the additional conditions: for all objects $C_1, C_2$ $\tau_{C_2, C_1} \circ \tau_{C_1, C_2} \cong 1_{C_1 \otimes C_2}$, and for all objects $C$, $\rho_C = \lambda_C \circ \tau_{C, e}: C \otimes e \cong C$. 
\end{definition}
An additional hexagon axiom is required to ensure that the $\tau$ natural isomorphism is compatible with $\alpha$.  The $\tau$ operator is called a ``swap" in Markov categories \citep{Fritz_2020}.  In most cases of interest in AI, the symmetric monoidal categories are {\em enriched} over some convenient base category ${\cal V}$, including vector spaces, or preorders such as the unit interval $[0,1]$, where the unique morphism from $a \rightarrow b$ exists if and only if $a \leq b$.
5
\begin{definition}
A {\bf {\cal V}-enriched category} consists of a regular category ${\cal C}$, such that for each pair of objects $x$ and $y$ in ${\cal C}$, the morphisms ${\cal C}(x,y) \in {\cal V}$, often referred to as a ${\cal V}$-hom object. For the case when $({cal V}, \leq, \otimes, 1)$ is a commutative monoidal preorder, we have the following conditions 
\begin{itemize}
    \item $1 \leq {\cal C}(x, x)$
    \item ${\cal C}(y,z) \otimes {\cal C}(x,y) \leq {\cal C}(x,z)$
\end{itemize}
\end{definition}

\subsection{LLM Markov Categories} 

In this section, we  introduce our framework of modeling LLMs by Markov categories \citep{Fritz_2020}, which have been proposed as a unifying categorical model for causal inference, probability and statistics. They are symmetric monoidal categories, which we reviewed above, combined with a comonoidal structure on each object. Importantly, Markov categories are semi-Cartesian because they do not use uniform copying, but contain a Cartesian subcategory defined by deterministic morphisms (see below).  We give a brief review of Markov categories, and significant additional details that are omitted can be found in \citep{Cho_2019,Fritz_2020,fritz:jmlr}.  The equations are written in the diagrammatic language of string diagrams, which can be shown to represent a formal language that is equivalent to writing down algebraic equations \citep{Selinger_2010}.

We can simply model the next-token distributions computed by an LLM as morphisms in a Markov category. We define the LLM Markov category more formally below. 

\begin{definition}
	\label{llm-markov_cat}
	A \emph{LLM Markov category} ${\cal C_{LLM}}$ \citep{Fritz_2020} is a symmetric monoidal category in which every object $X \in {\cal C_{LLM}}$ defines an LLM token, and is equipped with a commutative comonoid structure given by a comultiplication $\cop_X : X \to X \otimes X$ and a counit $\del_X : X \to I$, depicted in string diagrams as
	\beq
		\label{llm-cd}
		\begin{tikzpicture}
	\begin{pgfonlayer}{nodelayer}
		\node [style=bn] (0) at (7, 1) {};
		\node [style=bn] (1) at (-4, 0) {};
		\node [style=none] (2) at (7, -1) {};
		\node [style=none] (3) at (-4, -1) {};
		\node [style=none] (4) at (-5, 1) {};
		\node [style=none] (5) at (-3, 1) {};
		\node [style=none] (10) at (5, 0) {$=$};
		\node [style=none] (11) at (-7, 0) {$=$};
		\node [style=none] (12) at (3, 0) {$\del_X$};
		\node [style=none] (13) at (-9, 0) {$\cop_X$};
	\end{pgfonlayer}
	\begin{pgfonlayer}{edgelayer}
		\draw [bend right=45, looseness=1.25] (4.center) to (1);
		\draw (1) to (3.center);
		\draw [bend right=45] (1) to (5.center);
		\draw (2.center) to (0);
	\end{pgfonlayer}
\end{tikzpicture}

	\eeq
	and satisfying the commutative comonoid equations,
	\beq
		\label{llm_comonoid_ass}
		\begin{tikzpicture}
	\begin{pgfonlayer}{nodelayer}
		\node [style=none] (4) at (-3.25, -2) {};
		\node [style=none] (5) at (2.25, 0) {};
		\node [style=bn] (8) at (-3.25, -1) {};
		\node [style=none] (11) at (4.25, 2) {};
		\node [style=bn] (14) at (-2.25, 1) {};
		\node [style=none] (17) at (-4.25, 2) {};
		\node [style=none] (18) at (3.25, 2) {};
		\node [style=none] (20) at (-4.25, 0) {};
		\node [style=none] (22) at (2.25, 0) {};
		\node [style=none] (23) at (-3.25, 2) {};
		\node [style=none] (24) at (3.25, -2) {};
		\node [style=none] (28) at (-2.25, 0) {};
		\node [style=none] (30) at (4.25, 0) {};
		\node [style=none] (31) at (0, 0) {$=$};
		\node [style=bn] (32) at (2.25, 1) {};
		\node [style=none] (33) at (-2.25, 0) {};
		\node [style=bn] (34) at (3.25, -1) {};
		\node [style=none] (35) at (1.25, 2) {};
		\node [style=none] (36) at (-1.25, 2) {};
	\end{pgfonlayer}
	\begin{pgfonlayer}{edgelayer}
		\draw [style=none] (4.center) to (8);
		\draw [style=none, bend left=45] (8) to (20.center);
		\draw [style=none, bend right=45] (8) to (28.center);
		\draw [style=none] (33.center) to (14);
		\draw [style=none, bend left=45] (14) to (23.center);
		\draw [style=none, bend right=45] (14) to (36.center);
		\draw [style=none] (20.center) to (17.center);
		\draw [style=none] (24.center) to (34);
		\draw [style=none, bend left=45] (34) to (5.center);
		\draw [style=none, bend right=45] (34) to (30.center);
		\draw [style=none] (22.center) to (32);
		\draw [style=none, bend left=45] (32) to (35.center);
		\draw [style=none, bend right=45] (32) to (18.center);
		\draw [style=none] (30.center) to (11.center);
	\end{pgfonlayer}
\end{tikzpicture}

	\eeq
	\beq
		\label{llm_comonoid_other}
		\begin{tikzpicture}
	\begin{pgfonlayer}{nodelayer}
		\node [style=none] (0) at (-10, 1) {};
		\node [style=none] (1) at (-10, 2) {};
		\node [style=none] (2) at (-12, 1) {};
		\node [style=none] (3) at (-6.25, 0.5) {$=$};
		\node [style=none] (4) at (-3, 1) {};
		\node [style=none] (5) at (-11, -1) {};
		\node [style=bn] (6) at (-12, 1) {};
		\node [style=bn] (7) at (-4, 0) {};
		\node [style=none] (8) at (-5, 1) {};
		\node [style=bn] (9) at (-11, 0) {};
		\node [style=none] (10) at (-4, -1) {};
		\node [style=none] (11) at (-12, 1) {};
		\node [style=none] (12) at (-3, 1) {};
		\node [style=none] (13) at (-5, 2) {};
		\node [style=none] (14) at (-7.5, 2) {};
		\node [style=bn] (15) at (-3, 1) {};
		\node [style=none] (16) at (-8.75, 0.5) {$=$};
		\node [style=none] (17) at (-7.5, -1) {};
		\node [style=bn] (18) at (3.75, -0.25) {};
		\node [style=none] (19) at (2.75, 0.75) {};
		\node [style=none] (20) at (4.75, 0.75) {};
		\node [style=bn] (21) at (8.25, -0.25) {};
		\node [style=none] (22) at (7.25, 0.75) {};
		\node [style=none] (23) at (9.25, 0.75) {};
		\node [style=none] (24) at (6, 0) {$=$};
		\node [style=none] (25) at (3.75, -1.75) {};
		\node [style=none] (26) at (8.25, -1.75) {};
		\node [style=none] (27) at (2.75, 1.75) {};
		\node [style=none] (28) at (4.75, 1.75) {};
		\node [style=none] (29) at (7.25, 1.75) {};
		\node [style=none] (30) at (9.25, 1.75) {};
	\end{pgfonlayer}
	\begin{pgfonlayer}{edgelayer}
		\draw [style=none] (5.center) to (9);
		\draw [style=none, bend left=45] (9) to (11.center);
		\draw [style=none, bend right=45] (9) to (0.center);
		\draw [style=none] (6) to (2.center);
		\draw [style=none] (0.center) to (1.center);
		\draw [style=none] (10.center) to (7);
		\draw [style=none, bend left=45] (7) to (8.center);
		\draw [style=none, bend right=45] (7) to (12.center);
		\draw [style=none] (8.center) to (13.center);
		\draw [style=none] (15) to (4.center);
		\draw [style=none] (17.center) to (14.center);
		\draw [style=none, bend left=45] (18) to (19.center);
		\draw [style=none, bend right=45] (18) to (20.center);
		\draw [style=none, bend left=45] (21) to (22.center);
		\draw [style=none, bend right=45] (21) to (23.center);
		\draw (26.center) to (21);
		\draw (25.center) to (18);
		\draw (30.center) to (23.center);
		\draw (29.center) to (22.center);
		\draw [in=90, out=-90, looseness=0.75] (27.center) to (20.center);
		\draw [in=-90, out=90, looseness=0.75] (19.center) to (28.center);
	\end{pgfonlayer}
\end{tikzpicture}

	\eeq
	as well as compatibility with the monoidal structure,
	\begin{equation}
		\label{llm_delcopyAB}
		\resizebox{0.8\textwidth}{0.1\textwidth}{%
\begin{tikzpicture}
	\begin{pgfonlayer}{nodelayer}
		\node [style=none] (0) at (-7, 0) {$=$};
		\node [style=none] (1) at (2, 2.5) {$X\otimes Y$};
		\node [style=none] (2) at (11, -2) {};
		\node [style=bn] (3) at (11, -1) {};
		\node [style=none] (4) at (12, 0) {};
		\node [style=none] (5) at (12, 0) {};
		\node [style=none] (6) at (10, 0) {};
		\node [style=none] (7) at (15, -2) {};
		\node [style=bn] (8) at (15, -1) {};
		\node [style=none] (9) at (16, 0) {};
		\node [style=none] (10) at (16, 0) {};
		\node [style=none] (11) at (14, 0) {};
		\node [style=none] (12) at (10, 2) {};
		\node [style=none] (13) at (12, 2) {};
		\node [style=none] (14) at (14, 2) {};
		\node [style=none] (15) at (16, 2) {};
		\node [style=none] (16) at (-9, -1.5) {$X\otimes Y$};
		\node [style=none] (17) at (-5, -1.5) {$X$};
		\node [style=none] (18) at (8, 0) {$=$};
		\node [style=bn] (19) at (-9, 1) {};
		\node [style=bn] (20) at (-5, 1) {};
		\node [style=bn] (21) at (-3, 1) {};
		\node [style=bn] (22) at (4.5, -0.5) {};
		\node [style=none] (23) at (4.5, -2) {};
		\node [style=none] (24) at (3, 1) {};
		\node [style=none] (25) at (6, 1) {};
		\node [style=none] (26) at (-5, -1) {};
		\node [style=none] (27) at (-3, -1) {};
		\node [style=none] (28) at (-3, -1.5) {$Y$};
		\node [style=none] (29) at (-9, -1) {};
		\node [style=none] (30) at (6, 2.5) {$X\otimes Y$};
		\node [style=none] (31) at (4.5, -2.5) {$X\otimes Y$};
		\node [style=none] (32) at (3, 2) {};
		\node [style=none] (33) at (6, 2) {};
		\node [style=none] (34) at (10, 2.5) {$X$};
		\node [style=none] (35) at (12, 2.5) {$Y$};
		\node [style=none] (36) at (14, 2.5) {$X$};
		\node [style=none] (37) at (16, 2.5) {$Y$};
		\node [style=none] (38) at (11, -2.5) {$X$};
		\node [style=none] (39) at (15, -2.5) {$Y$};
	\end{pgfonlayer}
	\begin{pgfonlayer}{edgelayer}
		\draw [style=none] (2.center) to (3);
		\draw [style=none, bend left=45] (3) to (6.center);
		\draw [style=none, bend right=45] (3) to (5.center);
		\draw [style=none] (7.center) to (8);
		\draw [style=none, bend left=45] (8) to (11.center);
		\draw [style=none, bend right=45] (8) to (10.center);
		\draw [style=none] (12.center) to (6.center);
		\draw [style=none] (9.center) to (15.center);
		\draw [style=none, in=-90, out=90] (4.center) to (14.center);
		\draw [style=none, in=90, out=-90] (13.center) to (11.center);
		\draw (29.center) to (19);
		\draw (26.center) to (20);
		\draw (27.center) to (21);
		\draw (23.center) to (22);
		\draw [bend right=315] (22) to (24.center);
		\draw [bend right=45] (22) to (25.center);
		\draw (33.center) to (25.center);
		\draw (24.center) to (32.center);
	\end{pgfonlayer}
\end{tikzpicture}
}

	\end{equation}
	and naturality of $\del$, which means that
	\begin{equation}
		\label{llm_counit_nat}
		\begin{tikzpicture}
	\begin{pgfonlayer}{nodelayer}
		\node [style=bn] (0) at (-1.5, 2.5) {};
		\node [style=none] (8) at (-1.5, -1.5) {};
		\node [style=bn] (9) at (1.5, 1.5) {};
		\node [style=none] (10) at (1.5, -1.5) {};
		\node [style=none] (11) at (0.5, 0) {$=$};
		\node [style=morphism] (12) at (-1.5, 0) {$f$};
	\end{pgfonlayer}
	\begin{pgfonlayer}{edgelayer}
		\draw [style=none] (9) to (10.center);
		\draw (8.center) to (12);
		\draw (12) to (0);
	\end{pgfonlayer}
\end{tikzpicture}

	\end{equation}
	for every morphism $f$.
\end{definition}

Let us briefly explain these definitions. The $\mbox{del}_X: X \rightarrow I$ is essentially like integrating over a probability distribution, which always yields $1$. Hence, $I$, the unit of the tensor product, is the terminal object in affine CDU and Markov categories. Bayes rule turns into a {\em disintegration rule} \citep{Cho_2019}, which is only available in Markov categories with conditionals (i.e., where one can categorically refine $P(y | x)$ conditional distributions). Note that in the continuous case of random variables defined as measurable functions on real numbers, one has to take considerable care in defining conditioning \citep{halmos:book}. The $\mbox{copy}_X$ procedure is uniform, and deterministic, meaning if you take the tensor product of two variables $X \otimes Y$ and then copy the resulting object, that's exactly the same as first copying $X$ into $X \otimes X$ and $Y$ into $Y \otimes Y$, and then taking the tensor product, along with a swap operation (see Equation~\ref{llm_delcopyAB}). Only $\mbox{del}_X$ acts ``uniformly", meaning that if you process a variable $X$ using some function $f$ and then delete $f(X)$ (meaning marginalize it), that's equivalent to simply deleting $X$. However, $\mbox{copy}_X$ is not defined this way, and we discuss that subtlety below, as it will be important in understanding why Markov categories are semi-Cartesian. To convert them into a topos, we need the result to be Cartesian closed, which is why we need to use the Yoneda Lemma to construct the category of presheaves to guarantee obtaining a topos. 

\subsection{Cartesian Structure in Markov Categories}

\label{fox}

We now discuss a subcategory of Cartesian categories within  Markov that involves uniform $\mbox{copy}_X$ and $\mbox{del}_X$ morphisms. One fundamental property of Markov categories is that they are {\em semi-Cartesian}, as the unit object is also a terminal object.  But, a subtlety arises in how these copy and delete operators are modeled, as we discuss below. 

\begin{definition}
    A symmetric monoidal category ${\cal C}$ is {\bf Cartesian} if the tensor product $\otimes$ is the categorical product. 
\end{definition}

If ${\cal C}$ and ${\cal D}$ are symmetric monoidal categories, then a functor $F: {\cal C} \rightarrow {\cal D}$ is monoidal if the tensor product is preserved up to coherent natural isomorphisms. $F$ is strictly monoidal if all the monoidal structures are preserved exactly, including $\otimes$, unit object $I$, symmetry, associative and unit natural isomorphisms. Denote the category of symmetric monoidal categories with strict functors as arrows as {\bf MON}. Let us review the basic definitions given by \citet{Heunen2019}, which will give some further clarity on the Cartesian structure in affine CDU and Markov categories. 

\begin{definition}
 The subcategory of comonoids {\bf coMON} in the ambient category {\bf MON} of all symmetric monoidal categories is defined for any specific category {\cal C} as a collection of ``coalgebraic" objects $(X, \mbox{copy}_X, \mbox{del}_X)$, where $X$ is in {\cal C}, and arrows defined as comonoid homomorphisms  from $(X, \mbox{copy}_X, \mbox{del}_X)$ to  $(Y, \mbox{copy}_Y, \mbox{del}_Y)$ that act uniformly, in the sense that if $f: X \rightarrow Y$ is any morphism in {\cal C}, then: 

 \begin{eqnarray*}
     (f \otimes f) \circ \mbox{copy}_X = \mbox{copy}_Y \circ f \\
     \mbox{del}_Y \circ f = \mbox{del}_X 
 \end{eqnarray*}
\end{definition}

\citet{Heunen2019} define the process of ``uniform copying and deleting" in the category {\bf coMON}, which we now relate to Markov categories. A subtle difference worth emphasizing with Definition~\ref{llm-markov_cat} is that in Markov categories, only $\mbox{del}_X$ is ``uniform", but not $\mbox{copy}_X$ in the sense defined by \citet{Heunen2019}.  This distinction can be modeled in a cPROP category that is semi-Cartesian like Markov categories by suitably modifying the definition of the associated PROP map for copying. 

\begin{definition} \citep{Heunen2019}
    A symmetric monoidal category {\cal C} admits {\bf uniform deleting} if there is a natural transformation $e_X: X \xrightarrow{e_X} I$ for all objects in the  subcategory ${\cal C}_{\bf coMON}$ of comonoidal objects, where $e_I = \mbox{id}_I$, as shown in Equation~\ref{llm_counit_nat}. 
\end{definition}

This condition was referred to by \citet{Cho_2019} as a {\em causality} condition on the arrow $e_X$.  Essentially, it states that if you process some object and then discard it, it's equivalent to discarding it without processing. 

\begin{theorem} \citep{Heunen2019}
A symmetric monoidal category {\cal C} has uniform deleting if and only if $I$ is terminal. 
\end{theorem}

This property holds for Markov categories, as noted in \citep{Fritz_2020}, and a simple diagram chasing proof is given in \citep{Heunen2019}. 

\begin{definition}\citep{Heunen2019}
    A symmetric monoidal category {\cal C} has {\bf uniform copying} if there is a natural transformation $\mbox{copy}_X: X \rightarrow X \otimes X$ such that $\mbox{del}_I = \rho^{-1}_I$ satisfying Equation~\ref{llm_comonoid_ass} and Equation~\ref{llm_comonoid_other}. 
\end{definition}

We can now state an important result proved in \citep{Heunen2019} (Theorem 4.28), which relates to the more general results  shown earlier by \citet{fox}.  

\begin{theorem}\citep{fox,Heunen2019}
    The following conditions are equivalent for a symmetric monoidal category {\cal C}. 

    \begin{itemize} 

    \item The category {\cal C} is {\bf Cartesian} with tensor products $\otimes$ given by the categorical product and the tensor unit is given by the terminal object. 

    \item The symmetric monoidal category {\cal C} has {\bf uniform copying and deleting}, and Equation~\ref{llm_comonoid_ass} holds. 

    \end{itemize} 
\end{theorem}

As noted by \citet{Fritz_2020}, not all Markov categories are Cartesian, because their {\bf copy}$_X$ is not uniform, but only {\bf del}$_X$ is. For example, consider the  category {\bf FinStoch}, where a joint distribution is specified by the morphism $\psi: I \rightarrow X \otimes Y$. In this case, the marginal distributions can be formed as the composite morphisms
\begin{eqnarray*}
    I \xrightarrow{\psi} X \otimes  Y \xrightarrow{\mbox{del}_Y} X \\ 
    I \xrightarrow{\psi} X \otimes  Y \xrightarrow{\mbox{del}_X} Y \\ 
\end{eqnarray*}
But to require that in this case $\otimes$ is the categorical product implies that the marginal distributions defined as the above composites must be in bijection with the joint distribution.

\subsection{Quantum NLP with Monoidal Categories}

We briefly review previous work in modeling language with symmetric monoidal categories \citep{coecke2020mathematics,coecke2010mathematicalfoundationscompositionaldistributional}, combined with quantum computation \citep{Coecke_Kissinger_2017}.  This approach has been implemented in the {\tt DiscoPY} Python package  called {\tt Lambeq} \citep{kartsaklis2021lambeqefficienthighlevelpython}. A significant difference with our work is that the issue of categorical homotopy is not addressed in their work. The approach builds on a compositional theory of language defined by Lambek pregroup grammars, a type of category defined by a partially ordered monoid \citep{lambek}. 

In the Python package {\tt lambeq}, a natural language sentence is parsed using a statistical CCG parser \citep{yoshikawa-etal-2017-ccg}. Here, CCG denotes Combinatory Categorial Grammar \citep{lewis-steedman-2014-ccg}. Given a sentence $x$ of length $N$, the probability of a CCG tree ${\cal \tau}$ is computed as the product of the probabilities of the supertag (or categories) $c_i$, which are essentially factored models: 

\[ P({\cal \tau} | x) = \prod_{i \in [1, N]} P_{\mbox{tag}}(c_i | x) \]

There is an efficient parsing method called A$^*$ parsing \citep{klein-manning-2003-parsing} that can be used to implement parsing with CCG. {\tt lambeq} converts the CCG parse tree into a string diagram \citep{Selinger_2010}, which is a diagrammatic way to define the compositional structure of sentences. The string diagram, after some pre-processing, is then converted into a quantum circuit. There is a lot of interesting ideas in this body of work, but it is beyond the scope of this paper to do a more detailed discussion. The main takeaway message is that the issues of homotopy, which we discuss next, are not addressed here. An interesting future research question would be to combine the model category homotopy framework within a package like {\tt lambeq}. 

\section{Categorical Homotopy Theory} 

Before we introduce the framework of {\em model categories} \citep{Quillen:1967,Hovey2020ModelC} in this section, we need to give some background on homotopy theory \citep{Quillen:1967}. To reiterate our original problem, if we view sentence fragments as objects in an LLM Markov category, we run into the immediate problem that many of these objects should be treated as the same, even though they are non-isomorphic. Homotopy arose in the study of topology as a means to understand when two shapes are roughly the same (i.e., under homotopy). We explain the idea of capturing this using lifting diagrams. 

\subsection{Lifting Problems in  Categories}
\label{lift} 

 Lifting problems provide elegant ways to define basic notions in a wide variety of areas in mathematics \citep{lifting}.  For example, the notion of injective and surjective functions, the notion of separation in topology, and many other basic constructs can be formulated as solutions to lifting problems. Database queries in relational databases using languages like SQL can be formalized as lifting problems \citep{SPIVAK_2013}. Lifting problems define ways of decomposing structures into simpler pieces, and putting them back together again. We want to understand LLMs more abstractly in terms of lifting diagram and their application to model categories and simplicial sets.  
 
 \begin{definition}
 Let ${\cal C}$ be a  category. A {\bf {lifting problem}} in ${\cal C}$ is a commutative diagram $\sigma$ in ${\cal C}$. 
 \begin{center}
 \begin{tikzcd}
  A \arrow{d}{f} \arrow{r}{\mu}
    & X \arrow[]{d}{p} \\
  B  \arrow[]{r}[]{\nu}
&Y \end{tikzcd}
 \end{center} 
 \end{definition}
 
 \begin{definition}
 Let ${\cal C}$ be a  category. A {\bf {solution to a lifting problem}} in ${\cal C}$ is a morphism $h: B \rightarrow X$ in ${\cal C}$ satisfying $p \circ h = \nu$ and $h \circ f = \mu$ as indicated in the diagram below. 
 \begin{center}
 \begin{tikzcd}
  A \arrow{d}{f} \arrow{r}{\mu}
    & X \arrow[]{d}{p} \\
  B \arrow[ur,dashed, "h"] \arrow[]{r}[]{\nu}
&Y \end{tikzcd}
 \end{center} 
 \end{definition}

 \begin{definition}
 Let ${\cal C}$ be a category. If we are given two morphisms $f: A \rightarrow B$ and $p: X \rightarrow Y$ in ${\cal C}$, we say that $f$ has the {\bf {left lifting property}} with respect to $p$, or that p has the {\bf {right lifting property}} with respect to f if for every pair of morphisms $\mu: A \rightarrow X$ and $\nu: B \rightarrow Y$ satisfying the equations $p \circ \mu = \nu \circ f$, the associated lifting problem indicated in the diagram below. 
 \begin{center}
 \begin{tikzcd}
  A \arrow{d}{f} \arrow{r}{\mu}
    & X \arrow[]{d}{p} \\
  B \arrow[ur,dashed, "h"] \arrow[]{r}[]{\nu}
&Y \end{tikzcd}
 \end{center} 
admits a solution given by the map $h: B \rightarrow X$ satisfying $p \circ h = \nu$ and $h \circ f = \mu$. 
 \end{definition}


\subsection{LLM Category of Fractions}
The ``category of fractions" \citep{gabriel1967calculus} is intended to formalize the idea that there are weak equivalences in an LLM category -- sentences that mean the same thing -- that should be turned into isomorphisms in a constructed homotopy category. The problem of converting such weak equivalences into another homotopic category where they turn into isomorphisms is formalized using the following universal property \citep{borceux_1994}: 

\begin{definition}
Consider an LLM category ${\cal L}$ and a class $\Sigma$ of arrows of ${\cal L}$. The {\bf {category of fractions}} ${\cal L}(\Sigma^{-1})$ is said to exist when a category ${\cal L}(\Sigma^{-1})$ and a functor $\phi: {\cal L} \rightarrow {\cal L}(\Sigma^{-1})$ can be found with the following universal property: 

\begin{enumerate}
    \item $\forall f \in \Sigma, \phi(f)$ is an isomorphism. 
    \item If ${\cal L}'$ is a LLM category, and $F: {\cal L} \rightarrow {\cal L}'$ is a functor such that for all morphisms $f \in \Sigma$, $F(f)$ is an isomorphism, then there exists a unique functor $G: {\cal L}(\Sigma^{-1}) \rightarrow {\cal L}'$ such that $G \circ \phi = F$. 
\end{enumerate}
\end{definition}

 \section{Homotopy in LLM Categories}
 \label{homotopy}
 \begin{definition}
  Let  $X$ and $Y$ be two  objects in an LLM Markov category and suppose we are given a pair of morphisms $f_0, f_1: X \rightarrow Y$. A {\bf {homotopy}} from $f_0$ to $f_1$ is a morphism $h: \Delta^1 \times X \rightarrow Y$ satisfying $f_0 = h |_{{0} \times X}$ and $f_1 = h_{ 1 \times X}$. 
 \end{definition}

We now introduce a formal way to define algebraic invariants of a topological space defined by LLMs using the {\em nerve} of a category. As shown in \citep{segal}, the nerve of a category is a full and faithful embedding of a category as a simplicial object. In terms of simplicial sets, the nerve is simply the collection of $n$-simplices defined by composable pairs of arrows of length $n \geq 0$. 

\begin{definition}
The {\bf {classifying space}} ${\cal B L}$ of an LLM Markov category ${\cal L}$ is the topological space associated with the nerve of the category ${\cal L}$. 
\end{definition}
We can now define the equivalence classes of natural language sentences in a more abstract manner using abstract homotopy.  Two objects $C$ and $C'$ are in the same equivalence class ${\cal E}$ in an LLM category ${\cal L}$ if they related by a zig-zag  structure as shown in Figure~\ref{fig:llm-path}. 
\begin{figure*}[t]
    \centering
\[\begin{tikzcd}
	&& {C_1} &&& {C_2} &&&&& {C_{N}} \\
	\\
	C &&& {C'_1} &&& {C'_2} & \ldots & {C'_{N-1}} &&& {C'}
	\arrow[from=1-3, to=3-1]
	\arrow[from=1-3, to=3-4]
	\arrow[from=1-6, to=3-4]
	\arrow[from=1-6, to=3-7]
	\arrow[from=1-11, to=3-9]
	\arrow[from=1-11, to=3-12]
\end{tikzcd}\]
    \caption{Homotopy theory models equivalence classes in an LLM category. }
    \label{fig:llm-path}
\end{figure*}
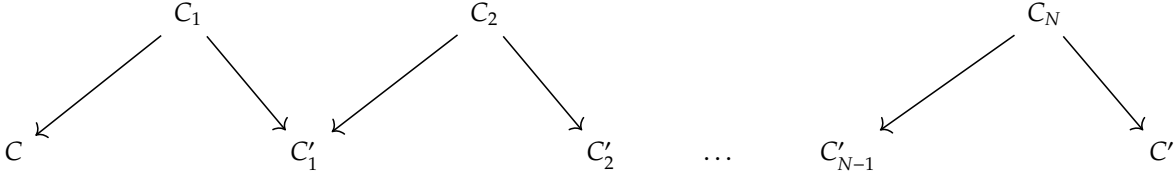
\begin{definition}
    Define the set of {\bf path components} of an LLM category ${\cal L}$ as the set of equivalence classes of the morphism relation on the objects by $\pi_0 {\cal L}$ (where $\pi_0$ usually refers to the ``fundamental group" of a category or a topological space, and is one of a series of algebraic invariants). 
\end{definition}
\begin{theorem}
\label{bc-connected}
The set of path components of the topological space ${\cal B L}$, namely $\pi_0 {\cal B L}$ is in bijection with the set of path components of an LLM category ${\cal L}$. 
\end{theorem}
{\bf Proof:} This is a special case of the result on path components of a category (see \citep{richter2020categories}). \qed 

This relationship between an LLM category ${\cal L}$ and its topological realization ${\cal BL}$ now gives us a homotopic characterization of LLMs that we desired in Figure~\ref{fig:llm-lifting}.  
We can define the higher homotopy groups of an LLM  category ${\cal L}$ as follows. 
\begin{definition}
    Given an LLM Markov category ${\cal L}$, and an arbitrary object $X$ in ${\cal C}$. For $n \geq 0$, the $n^{th}$ homotopy group of ${\cal C}$ with respect to the basepoint $X$ is defined as 
    \[ \pi_n(X, {\cal L}) = \pi_n({\cal BL}, [X]) \]
    where $[X]$ is the $0$-simplex associated to the basepoint $X$. 
\end{definition}
We know from Definition~\ref{llm-syntax} or Definition~\ref{llm-semantics-defn} that an LLM category ${\cal L}$ is enriched over a symmetric monoidal category, there is a wealth of known results that can be brought to bear on its homotopic structure. 
\begin{definition}
An $H$-space associated with an LLM is a topological space $X$ with a chosen base point $x_0$, and a continuous map $\mu: X \times X \rightarrow X$ such that the maps $\mu(x_0,.)$ and $\mu(., x_0)$ are homotopic to the identity map on $X$ with respect to homotopies that preserve the basepoint $x_0$. An LLM $H$-space is associative if $\mu$ is associative up to homotopy, and it is commutative if $\mu$ is commutative up to homotopy. 
\end{definition}
\begin{definition}
An LLM $H$-space $X$ is group-like if there is a continuous map $\chi: X \rightarrow X$ such that $\mu \circ (\mbox{1}_{id} \times \chi) \circ \Delta$ is homotopic to the identity, where $\Delta$ is the the diagonal map on $X$. 
\end{definition}
\begin{theorem}
Let ${\cal L}$ be an LLM Markov category. Then its classifying space ${\cal BL}$ is an associative and commutative $H$-space. 
\end{theorem}
{\bf Proof:} The proof of this theorem is based on a simple diagram chase, building on the standard result for (small) symmetric monoidal categories. 
\begin{center}
\[\begin{tikzcd}
	{{\cal BL} \times {\cal BL}} &&& {{\cal B}(L \times L)} \\
	\\
	&&& {{\cal BL}}
	\arrow["\cong", from=1-1, to=1-4]
	\arrow["\mu"', from=1-1, to=3-4]
	\arrow["{{\cal B}(\otimes)}", from=1-4, to=3-4]
\end{tikzcd}\]    
\end{center} \qed 

\subsection{LLM Algebraic K-Theory}
\label{ktheory}
Finally, we build on the above results to define  LLM homotopy using the Grayson-Quillen K-Theory \citep{quillen,grayson}. 
\begin{definition}
    Let $({\cal L}, \otimes, e, \tau)$ be an LLM Markov category. Denote by ${\cal L}^{-1} {\cal L}$ the category whose objects are pairs of objects of ${\cal L}$. Morphisms in ${\cal L}^{-1} {\cal L}$ from $(C_1, D_1)$ to $(C_2, D_2)$ are equivalence classes of pairs of morphisms 

    \[(f: C_1 \otimes E \rightarrow C_2, g: D_1 \otimes E \rightarrow D_2) \]

    where $E$ is an object of ${\cal L}$. Such pairs of morphisms are equivalent to 

    \[(f': C'_1 \otimes E' \rightarrow C_2, g': D_1 \otimes E' \rightarrow D_2) \]

    if there is an isomorphism $h \in {\cal L}(E, E')$ such that the following diagram commutes: 

\begin{small}
\begin{center}
\begin{tikzcd}
	{(C_1 \otimes E, D_1 \otimes E)} &&& {(C_1 \otimes E', D_1 \otimes E')} \\
	\\
	&& {(C_2, D_2)}
	\arrow["{(1_{C_1} \otimes h, 1_{D_1} \times h)}", from=1-1, to=1-4]
	\arrow["{(f, g)}"', from=1-1, to=3-3]
	\arrow["{(f', g')}", from=1-4, to=3-3]
\end{tikzcd}
\end{center}
\end{small}

The category ${\cal L}^{-1} {\cal L}$ is called the {\bf Grayson-Quillen} construction of ${\cal L}$. 

\end{definition}
%
%
\begin{theorem}
    The category ${\cal L}^{-1} {\cal L}$ is symmetric monoidal as well, and there is a lax monoidal functor $j: {\cal L} \rightarrow {\cal L}^{-1} {\cal L}$, and $\pi_0({\cal B L}^{-1} {\cal L})$ is an Abelian group. 
\end{theorem}
{\bf Proof:} The proof follows readily from the general result in \citep{richter2020categories}, Lemma 13.3.2. \qed 
\begin{definition}
    The {\bf K-Theory} space associated with an LLM category ${\cal L}$ is the classifying space ${\cal K L} = {\cal B L}^{-1} {\cal L}$, where the $n^{th}$ K-group of ${\cal L}$ is its $n^{th}$ homotopy group $\pi_n {\cal B L}^{-1} {\cal L}$.  Specifically, the fundamental group $\pi_0({\cal B L}^{-1} {\cal L})$ is the Grothendieck group completion of the Abelian monoid induced by the path components $\pi_0({\cal L})$.   
\end{definition}
Let us now connect this procedure with  the $0^{th}$ homotopy group of the LLM equivalence class.  
\begin{theorem}
The higher algebraic K-theory of an LLM category ${\cal L}$ is defined as
\[ K_0({\cal L}) = \pi_0({\cal KL}) \simeq G_0 (\pi_0 ({\cal L}))  \]
\end{theorem}
{\bf Proof:} The proof follows readily from the more general result that holds in any symmetric monoidal category (see Lemma 13.3.4 in \citep{richter2020categories}). \qed 
\subsection{LLM Groupoids}
A groupoid is a category whose  morphisms are invertible. We can characterize LLM equivalence classes in terms of the classifying space of their induced groupoids. 
\begin{definition}
Define the LLM groupoid as the category ${\cal L}_{G}$ whose objects are defined as  the equivalence classes of the connected paths of the category, and whose invertible morphisms correspond  to invertible edges that map from an equivalence class back to itself.  
\end{definition}
%

%
\begin{theorem}
The classifying space of the LLM groupoid category ${\cal B L}_{G}$ is defined as 
\[ {\cal B L}_{G} = \bigsqcup_{i} {\cal B L}^i_{G} \]
where disjoint sum  index $i$ ranges over equivalence classes. 
\end{theorem}
{\bf Proof:} The proof follows readily from the general result in \citep{richter2020categories} on classifying spaces of groupoids. \qed 
%
%

\section{LLM as a Model Category} 
\label{sSets}
To define a model category over an LLM Markov category, we need to convert it into a simplicial set using the nerve functor \citep{segal}. There are number of tricky issues here to resolve, which lie beyond the scope of this introductory paper. For one, the standard nerve functor applied to a category produces a simplicial set (intuitively, each $n$-simplex is the set of all $n$-length composable morphisms), but this special type of simplicial set is a {\em quasi-category} \citep{quasicats}. Such quasi-categories only allow inner horns to be filled, and do not in general lead to model categories. However, and this is quite crucial to our analysis, when we have objects in an LLM Markov category that induce a groupoid structure because they are essentially paraphrases of each other that should reflect the same next-token distribution, then the nerve functor does yield a model category \citep{riehl2019infinitycategorytheoryscratch}. It is also the case that for {\em permutative categories}, the nerve functor for symmetric monoidal categories is well-behaved \citep{richter2020categories}. There are subtleties here that we are glossing over, but they are not essential to the main objective of this paper. 

\subsection{Simplicial Sets} 
\begin{figure}[t]
    \centering
    \includegraphics[width=0.4\linewidth]{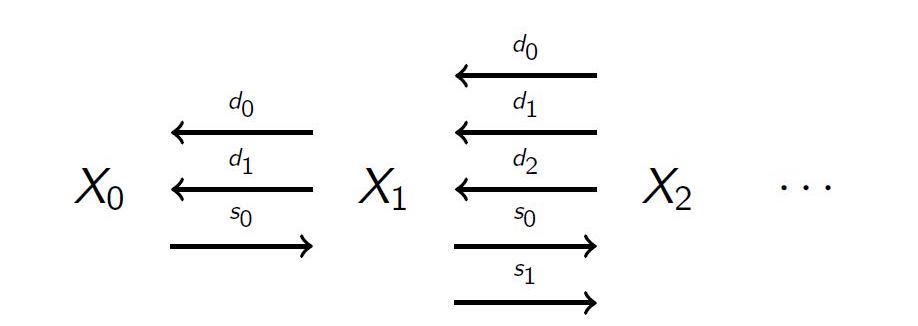}
    \caption{A simplicial set \citep{may1992simplicial} representation of an LLM consists of a graded sequence of sets, where the $0$-simplex is a set of tokens, the $1$-simplex is a set of pairs of adjacent tokens, and in general, the $n$-simplex is a set of sequences of tokens of length $n$. Face and degeneracy operators map between an $n$-simplex and its lower or higher dimensional simplices.}
    \label{fig:llm-ssets}
\end{figure}
The central goal of this section is to show that LLMs define model categories (Theorem~\ref{llm-model-category}). To show this result, we need to introduce the categorical machinery of simplicial sets \citep{may1992simplicial}, a combinatorial representation of topological spaces that are CW-complexes \citep{munkres:algtop}. Simplicial sets generalize directed graphs and form the foundation of higher-order category theory \citep{weakkan,quasicats,Lurie:higher-topos-theory}. 

The category $\Delta$ has non-empty ordinals $[n] = \{0, 1, \ldots, n\}$ as objects, and order-preserving maps $[m] \rightarrow [n]$ as arrows. An important property in $\Delta$ is that any arrow is decomposable as a composition of an injective and a surjective mapping,  each of which is decomposable into a sequence of elementary injections $\delta_i: [n] \rightarrow [n+1]$, called {\em {coface}} mappings, which omits $i \in [n]$, and a sequence of elementary surjections $\sigma_i: [n] \rightarrow [n-1]$, called {\em {co-degeneracy}} mappings, which repeats $i \in [n]$. The fundamental simplex $\Delta([n])$ is the presheaf of all morphisms into $[n]$, that is, the representable functor $\Delta(-, [n])$.  The Yoneda Lemma \citep{maclane:71}  assures us that an $n$-simplex $x \in X_n$ can be identified with the corresponding map $\Delta[n] \rightarrow X$. Every morphism $f: [n] \rightarrow [m]$ in $\Delta$ is functorially mapped to the map $\Delta[m] \rightarrow \Delta[n]$ in ${\cal S}$. 

Any morphism in the category $\Delta$ can be defined as a sequence of {\em co-degeneracy} and {\em co-face} operators, where the co-face operator $\delta_i: [n-1] \rightarrow [n], 0 \leq i \leq n$ is defined as: 
\[ 
\delta_i (j)  =
\left\{
	\begin{array}{ll}
		j,  & \mbox{for } \ 0 \leq j \leq i-1 \\
		j+1 & \mbox{for } \  i \leq j \leq n-1 
	\end{array}
\right. \] 

Analogously, the co-degeneracy operator $\sigma_j: [n+1] \rightarrow [n]$ is defined as 
\[ 
\sigma_j (k)  =
\left\{
	\begin{array}{ll}
		j,  & \mbox{for } \ 0 \leq k \leq j \\
		k-1 & \mbox{for } \  j < k \leq n+1 
	\end{array}
\right. \] 

Note that under the contravariant mappings, co-face mappings turn into face mappings, and co-degeneracy mappings turn into degeneracy mappings. That is, for any simplicial object (or set) $X_n$, we have $X(\delta_i) \coloneqq d_i: X_n \rightarrow X_{n-1}$, and likewise, $X(\sigma_j) \coloneqq s_j: X_{n-1} \rightarrow X_n$. 

The compositions of these arrows define certain well-known properties \citep{may1992simplicial,richter2020categories}: 
\begin{eqnarray*}
    \delta_j \circ \delta_i &=& \delta_i \circ \delta_{j-1}, \ \ i < j \\
    \sigma_j \circ \sigma_i &=& \sigma_i \circ \sigma_{j+1}, \ \ i \leq j \\ 
    \sigma_j \circ \delta_i (j)  &=&
\left\{
	\begin{array}{ll}
		\sigma_i \circ \sigma_{j+1},  & \mbox{for } \ i < j \\
		1_{[n]} & \mbox{for } \  i = j, j+1 \\ 
		\sigma_{i-1} \circ \sigma_j, \mbox{for} \ i > j + 1
	\end{array}
\right.
\end{eqnarray*}

\begin{definition}
    A {\bf simplicial object} is a contravariant functor $X: \Delta^{op} \rightarrow {\cal C}$, where $\Delta$ is the category of ordinals introduced above, and ${\cal C}$ is any category of objects and arrows. 
\end{definition}

Simplicial  sets are contravariant functors from $\Delta^{op}$ into the category of ${\bf Sets}$. We are generalizing this notion to any category, including LLM categories defined earlier in Section~\ref{llmcat-defns}.  Each $n$-simplex $X[n]$ is in effect a collection of objects of the category ${\cal C}$, and the morphisms are those that are mapped from the weakly order-preserving morphisms of $\Delta$.

\begin{example}
The ``vertices'' of a simplicial object $X$ in a category ${\cal C}$ are the objects $X_0$ in  ${\cal C}$, and the ``edges'' $X_1$ are its arrows $f: C_i \rightarrow C_j$, where $C_i$ and $C_j$ are objects in ${\cal C}$. Note that $X_0$ is a contravariant functor $X: [0] \rightarrow {\cal C}$, and since $[0]$ has only one object, the effect of this functor is to pick out objects in ${\cal C}$. The simplicial object $X_1: [1] \rightarrow {\cal C}$. Given any such arrow, the face operators $d_0 f = C_j$ and $d_1 f = C_i$ recover the source and target of each arrow. Also, given an object $X$ of category ${\cal C}$, we can regard the degeneracy operator $s_0 X$ as its identity morphism ${\bf 1}_X: X \rightarrow X$. 
\end{example}

\begin{example} 
Given a  category ${\cal C}$, we can define an $n$-simplex $X_n$ of a simplicial object in ${\cal C}$ with \mbox{the sequence: }
\[ X_n = C_o \xrightarrow[]{f_1} C_1 \xrightarrow[]{f_2} \ldots \xrightarrow[]{f_n} C_n \] 
the face operator $d_0$ applied to $X_n$ yields the sequence 
\[ d_0 X_n = C_1 \xrightarrow[]{f_2} C_2 \xrightarrow[]{f_3} \ldots \xrightarrow[]{f_n} C_n \] 
where the object $C_0$ is ``deleted'' along with the morphism $f_0$ leaving it. 

\end{example} 

\begin{example} 
Given a category ${\cal C}$, and an $n$-simplex $X_n$ of the simplicial object in a category  $X_n$, the face operator $d_n$ applied to $X_n$ yields the sequence 
\[ d_n X_n = C_0 \xrightarrow[]{f_1} C_1 \xrightarrow[]{f_2} \ldots \xrightarrow[]{f_{n-1}} C_{n-1} \] 
where the object $C_n$ is ``deleted'' along with the morphism $f_n$ entering it. 

\end{example} 

\begin{example} 
Given a category  ${\cal C}$, and an $n$-simplex $X_n$ of the simplicial object, 
the face operator $d_i, 0 < i < n$ applied to $X_n$ yields the sequence 
\[ d_i X_n = C_0 \xrightarrow[]{f_1} C_1 \xrightarrow[]{f_2} \ldots C_{i-1} \xrightarrow[]{f_{i+1} \circ f_i} C_{i+1} \ldots \xrightarrow[]{f_{n}} C_{n} \] 
where the object $C_i$ is ``deleted'' and the morphisms $f_i$ is composed with morphism $f_{i+1}$.  

\end{example} 

\begin{example} 
Given a category ${\cal C}$, and an $n$-simplex $X_n$ of the simplicial object defined over the  category, 
the degeneracy operator $s_i, 0 \leq i \leq n$ applied to $X_n$ yields the sequence 
\[ s_i X_n = C_0 \xrightarrow[]{f_1} C_1 \xrightarrow[]{f_2} \ldots C_{i} \xrightarrow[]{{\bf 1}_{C_i}} C_{i} \xrightarrow[]{f_{i+1}} C_{i+1}\ldots \xrightarrow[]{f_{n}} C_{n} \] 
where the object $C_i$ is ``repeated'' by inserting its identity morphism ${\bf 1}_{C_i}$. 

\end{example} 

\subsection{Kan Complexes}

To prove that LLMs define model categories, we use lifting diagrams to introduce {\em Kan complexes}.  
 
 \begin{definition}
 The {\bf {standard simplex}} $\Delta^n$ is the simplicial set defined by all mappings out of $[m]$: 
 \[ ([m] \in \Delta) \mapsto {\bf Hom}_\Delta([m], [n]) \] 
 
 By convention, $\Delta^{-1} \coloneqq \emptyset$. The standard $0$-simplex $\Delta^0$ maps each $[n] \in \Delta^{op}$ to the single element set $\{ \bullet \}$. 
 \end{definition}
 
 \begin{definition}
 Let $X$ denote a simplicial object, where $X_n$ is its $n^{th}$ simplex. If for every integer $n \geq 0$, we are given a subset $Y_n \subseteq X_n$, such that the face and degeneracy maps 
 \[ d_i: X_n \rightarrow X_{n-1} \ \ \ \ s_i: X_n \rightarrow X_{n+1} \] 
 applied to $Y_n$ result in 
 \[ d_i: Y_n \rightarrow Y_{n-1} \ \ \ \ s_i: Y_n \rightarrow Y_{n+1} \] 
 then the collection $\{ Y_n \}_{n \geq 0}$ defines a {\bf {simplicial subset}} $Y_\bullet \subseteq X_\bullet$
 \end{definition}
 
 \begin{definition}
 The {\bf {boundary}} is a simplicial set $(\partial \Delta^n): \Delta^{op} \rightarrow$ {\bf {Set}} defined as
 \[ (\partial \Delta^n)([m]) = \{ \alpha \in {\bf Hom}_\Delta([m], [n]): \alpha \ \mbox{is not surjective} \} \]
 \end{definition}
 
 Note that the boundary $\partial \Delta^n$ is a simplicial subset of the standard $n$-simplex $\Delta^n$. 
 
 \begin{definition}
 The {\bf {Horn}} $\Lambda^n_i: \Delta^{op} \rightarrow$ {\bf {Set}} is defined as
 \[ (\Lambda^n_i)([m]) = \{ \alpha \in {\bf Hom}_\Delta([m],[n]): [n] \not \subseteq \alpha([m]) \cup \{i \} \} \] 
 \end{definition}
 
 Intuitively, the Horn $\Lambda^n_i$ can be viewed as the simplicial subset that results from removing the interior of the $n$-simplex $\Delta^n$ together with the face opposite its $i$th vertex.   
 \begin{definition}
 Let $f: X \rightarrow S$ be a morphism of simplicial objects in a  category ${\cal C}$. We say $f$ is a {\bf Kan fibration} if, for each $n > 0$, and each $0 \leq i \leq n$, every lifting problem. 
 \begin{center}
 \begin{tikzcd}
  \Lambda^n_i \arrow{d}{} \arrow{r}{\sigma_0}
    & X \arrow[]{d}{f} \\
  \Delta^n \arrow[ur,dashed, "\sigma"] \arrow[]{r}[]{\bar{\sigma}}
&S \end{tikzcd}
 \end{center}  
 admits a solution. More precisely, for every map of simplicial sets $\sigma_0: \Lambda^n_i \rightarrow X$ and every $n$-simplex $\bar{\sigma}: \Delta^n \rightarrow S$ extending $f \circ \sigma_0$, we can extend $\sigma_0$ to an $n$-simplex $\sigma: \Delta^n \rightarrow X$ satisfying $f \circ \sigma = \bar{\sigma}$. 
 \end{definition}
 
 \begin{example}
Given a simplicial object $X$ in a  category ${\cal C}$, a projection map $X \rightarrow \Delta^0$ that is a Kan fibration is called a {\bf {Kan complex}}. 
\end{example} 

\begin{example}
Any isomorphism between simplicial objects in a  category ${\cal C}$ is a Kan fibration. 
\end{example}

\begin{example}
The collection of Kan fibrations in  categories is closed under retracts. 
\end{example}

\subsection{Model Categories}
\label{mc}
 An elegant framework for abstract homotopy was developed by \citet{Quillen:1967} called a {\em model category}, which is based on classifying the arrows into three types of morphisms: a set of {\em cofibrations} (e.g., the ``injective" arrow $i$ on the left side of Figure~\ref{fig:llm-lifting}), a set of {\em fibrations} (the surjective arrow $p$ on the right-hand side of Figure~\ref{fig:llm-lifting}), and a set of {\em weak equivalences}, which have to satisfy the following properties. One of our main results (see Theorem~\ref{llm-model-category}) is that LLMs define model categories. 
\begin{definition}
    A {\bf model category} ${\cal M}$ is a category equipped with three classes of morphisms: {\em weak equivalences}, {\em cofibrations} and {\em fibrations}. A map which is a (co)fibration and a weak equivalence is called an {\em acyclic (co)fibration}. A model category ${\cal M}$ with these three classes of morphisms must satisfy the following five conditions: 
    \begin{enumerate}
        \item {\bf MC1}: The category ${\cal M}$ has all small limits and colimits. 
        \item {\bf MC2}: If $f$ and $g$ are any two maps in ${\cal M}$ such that $g \circ f$ is defined, and if any two of the maps -- $f$, $g$ and $g \circ f$ -- are weak equivalences, so is the third. 
        \item {\bf MC3}: If $f$ and $g$ are maps in ${\cal M}$ such that $f$ is a {\em retract} of $g$, and $g$ is a weak equivalence, fibration, or cofibration, then so is $f$. 
        \item {\bf MC4}: Given a lifting diagram such as the one defined below, the dotted arrow $h$ exists if either $f$ is a cofibration and $p$ is an acyclic fibration, or $i$ is an acyclic cofibration and $p$ is a fibration. 
        \begin{center}
 \begin{tikzcd}
  A \arrow{d}{f} \arrow{r}{\mu}
    & X \arrow[]{d}{p} \\
  B \arrow[ur,dashed, "h"] \arrow[]{r}[]{\nu}
&Y \end{tikzcd}
 \end{center} 
 \item {\bf MC5}: Any map $g$ in ${\cal M}$ can be factored in two ways: $g = q \circ i$ where $i$ is a cofibration and $q$ is an acyclic fibration, or $g = p \circ j$ where $j$ is an acyclic cofibration and $p$ is  a fibration. 
    \end{enumerate}
\end{definition}
\begin{definition}
Let ${\cal C}$ be any category with a subcategory of weak equivalences ${\cal W}$. The ``free category"   $F({\cal C}, {\cal W}^{-1})$ is defined on the arrows of ${\cal C}$ by adding in the reversal of arrows in ${\cal W}$, which has the same objects as ${\cal C}$, and its morphisms are strings of composable morphisms $(f_1, \ldots, f_n)$ where $f_i$ is either an arrow of ${\cal C}$ or a reversal $w^{-1}_i$  of an arrow $w_i$ in ${\cal W}$. The {\bf homotopy category}  $\mbox{Ho} {\cal C}$ is defined as the quotient of the ``free" category by the relations $1_x = (1_x)$ for all objects $x$ in ${\cal C}$, $(f, g) = (g \circ f)$ for all composable morphisms $f, g \in {\cal C}$ and $1_{\mbox{dom}_w} = (w^{-1}, w)$ for all $w \in {\cal W}$.  
\end{definition}
A detailed overview of model categories is given in many textbooks \citep{Hovey2020ModelC,may2012more,Quillen:1967}. 
\subsection{LLM Model Categories}
A key result of our paper is that LLM Markov categories define model categories. We introduce two basic definitions before sketching out the proof. 
\begin{definition}
 Let $C$ and $C'$ be a pair of objects in an LLM Markov category ${\cal L}$. We say $C$ is {\bf {a retract}} of $C'$ if there exists maps $i: C \rightarrow C'$ and $r: C' \rightarrow C$ such that $r \circ i = \mbox{id}_{\cal C}$. 
 \end{definition}
 \begin{definition}
 Let ${\cal L}$ be an LLM Markov category. We say a morphism $f: C \rightarrow D$ is a {\bf {retract of another morphism}} $f': C \rightarrow D$ if it is a retract of $f'$ when viewed as an object of the functor category {\bf {Hom}}$([1], {\cal L})$. Recall that $[1] = \{0, 1\}$ has one nontrivial morphism $0 \rightarrow 1$.  A collection of morphisms $T$ of ${\cal L}$ is {\bf {closed under retracts}} if for every pair of morphisms $f, f'$ of ${\cal L}$, if $f$ is a retract of $f'$, and $f'$  is in $T$, then $f$ is also in $T$. 
 \end{definition}
\begin{theorem}
\label{llm-model-category}
    LLM Markov categories define model categories. 
\end{theorem}
 {\bf Proof Sketch:} The proof is a special case of the result that simplicial sets define model categories \citep{Quillen:1967,Hovey2020ModelC,may2012more}: 

 \begin{itemize}
     \item {\em Fibrations are Kan complexes}: Kan complexes are those that satisfy the lifting property for all horn inclusions. Simplicial sets constructed using the nerve functor over LLM Markov categories  are Kan complexes as each $n$-simplex is a sequence of $n$-length tokens. Any $n-1$ simplex $X_{n-1}$ of an LLM is a retract of an $n$-simplex $X_n$ of an LLM by applying a face operator (see Examples 2-4). 
     \item {\em Cofibrations are monomorphisms of simplicial sets}: If we define mappings between LLM simplicial sets in terms of the constituent mappings between their corresponding $n$-simplices (which are sets of composable morphisms), then cofibrations in LLM model categories are the injective mappings. 
     \item {\em Weak equivalences are associated with topological emebddings of simplicial sets}: The weak equivalence between two LLM simplicial sets are the corresponding weak equivalences between their topological realizations (we review topological embeddings of LLMs in Section~\ref{top-simp}). 
 \end{itemize} \qed 

 \section{Cartesian Topological Spaces}

\label{cartesian}

In this section, we discuss ``nice" categories of topological spaces, showing that the obvious categorization leads to a symmetric monoidal structure where the tensor product $\otimes$ is defined as a product of topological spaces, but it is not a Cartesian category. Restricting to the subcategory of CW complexes (or compactly generated weak Hausdorff spaces \citep{munkres:algtop}) allows defining a suitable Cartesian structure. This review will motivate the construction of topological structures from LLM categories  through the use of simplicial sets, which was discussed in the main paper. 

\subsection{Cartesian Structure in Topological Spaces}

Our goal in this paper is to construct topological representations of equivalence classes of natural language sentences as represented in an LLM. To this end, we need to understand some basics about categories of topological spaces \citep{borceux_1994}  and why the simplicial set construction described in the main paper leads to a nice topological structure. 

\begin{definition}
The category {\bf Top} defines the category of all topological spaces, where objects are topological spaces, and arrows are continuous functions from one space into another. 
\end{definition}

The category {\bf Top} has a natural symmetric monoidal structure, as defined in the following results (see \citep{borceux_1994}, Chapter 7). 

\begin{theorem}
    The category {\bf Top} has a natural closed symmetric monoidal structure imposed on it where

    \begin{itemize}
    \item The set underlying the tensor product $X \otimes Y$ of two spaces $X$ and $Y$ must necessarily be the Cartesian product $X \times Y$ of the underlying sets. 

    \item The set underlying the internal ``hom object" function space $[X, Y]$ is necessarily the set ${\bf Top}(X, Y)$ of continuous mappings from $X$ to $Y$. 
    \end{itemize} 
\end{theorem}

However, this natural monoidal structure does not have the right topological properties, because of the following result: 

\begin{theorem}
    The category {\bf Top} is not Cartesian closed. 
\end{theorem}

As before, we require that a Cartesian structure must require the ability to construct exponential objects. 

\begin{definition}
    A topological space $X$ is {\bf exponentiable} if the product functor $- \times X: {\bf Top} \rightarrow {\bf Top}$ has a right adjoint. 
\end{definition}

Fortunately, it is possible to construct such a subcategory of topological spaces. 

\begin{definition}
    A topological space $X$ is locally compact when every neighborhood of a point contains a compact neighborhood of this point. 
\end{definition}

This restriction leads to a satisfactory outcome for constructing exponentiable objects. 

\begin{theorem}
A locally compact space is exponentiable. The right adjoint to the functor $- \times X$ is given by the functor mapping $Y$ to the set ${\bf Top}(X,Y)$ of continuous functions, topologized with the compact open topology. 
\end{theorem}

The compact open topology on ${\bf Top}(X, Y)$ is defined as 

\[ \langle K, U \rangle = \{ f \in {\bf Top}(X, Y) | f(K) \subseteq U\]

where $K$ is over the compact subsets of $X$ and $U$ is over the open subsets of $Y$. 

An ideal Cartesian structure on topological spaces is given by the category of compact Hausdorff spaces with compact open mappings. 

\begin{theorem}
    The category {\bf CGHaus} of compact Hausdorff spaces and compactly continuous mappings is Cartesian closed. 
\end{theorem}

Finally, we note that the simplicial set construction described in the main paper yields CW complexes (compactly generated weak Hausdorff), which is defined as the following: 

\begin{definition}
A space $X$ is {\bf weak Hausdorff} if the image of any continuous map with compact Hausdorff domain is closed in $X$. 
\end{definition}

These properties will be useful in understanding the topological structure of classifying spaces of LLM categories, in particular they lead to associative and multiplicative spaces called $H$-spaces.

 \subsection{Topological Embeddings as Coends}

We now bring in the perspective that topological embeddings of simplicial objects can be interpreted as a coend \citep{loregian_2021,maclane:71} as well (we review (co)end calculus in Section~\ref{coends}): 
\[ F: \Delta^o \times \Delta \rightarrow \mbox{Top} \] 
where $F$ acts {\em contravariantly} as a functor from $\Delta$ to ${\bf Sets}$ mapping $[n] \mapsto X_n$,  and {\em covariantly} mapping $[m] \mapsto \Delta^m$ as a functor from $\Delta$ to the category ${\bf Top}$ of topological spaces.  The coend  defines a topological embedding of a simplicial object $X$,  where $X_n$ represents composable morphisms of length $n$ with an associated probability$p$. Given this simplicial object, we can now construct a topological realization of it as a coend object 
\[ \int^n (\mbox{X}_n) \cdot {\bf \Delta} n \] 
where $\mbox{X}: \Delta^{op} \rightarrow {\cal C}$ is the simplicial object defined by the contravariant functor from the simplicial category $\Delta$ into the category of simplicial objects, and ${\bf \Delta}: |\Delta | \rightarrow {\bf Top}$ is a functor from the topological $n$-simplex realization of the simplicial category $\Delta$ into topological spaces ${\bf Top}$. Here, the notation $(\mbox{X}_n) \cdot {\bf \Delta} n$ means ``co-power", and represents the disjoint union $\sqcup_{\mbox{X}_n} {\bf \Delta}$ of $\mbox{X}_n$ copies of ${\bf \Delta_n}$ the topological $n$-simplex.

\section{LLM Fuzzy Simplicial Objects}

\label{llm-lot}

Now we study ``fuzzy"  simplicial sets \citep{spivak:fuzzy,umap} defined over  the LLM categories in Section~\ref{llmcat-defns}. 
\begin{definition}
    A {\bf fuzzy simplicial object} is a contravariant functor $X: \Delta^{op} \times [0,1] \rightarrow {\cal C}$, where ${\cal C}$ is any category. Each object in $X$ is a fuzzy n-simplex is a pair $(X[n], p)$ where $p$ represents the probability or ``strength" of the $n$-simplex. An $n$-simplex has a strength at most the minimum of its faces. 
\end{definition}
\begin{definition}
    A {\bf An LLM fuzzy simplicial object} is a contravariant functor $X: \Delta^{op}\times [0,1] \rightarrow {\cal L}$, where ${\cal L}$ is a  category of LLMs. Each object in $X$ is a pair $(X[n], p)$, where $p \in [0,1]$  is the ``strength" associated with the $n$-length token sequences in $X[n]$. 
\end{definition}
%
\subsection{Topological Embedding}
\label{top-simp}
Simplicial objects can be embedded in a topological space using a construction originally proposed by \citet{milnor}. 

\begin{definition}
 The {\bf geometric realization} $|X|$ of a  simplicial object $X$ in an LLM category ${\cal L}$  defined as the topological space 

 \[ |X| = \bigsqcup_{n \geq 0} X_n \times { \bf \Delta}^n / ~\sim \]

 where the $n$-simplex $X_n$ is assumed to have a {\em discrete} topology (i.e., all subsets of $X_n$ are open sets), and ${\bf \Delta}^n$ denotes the {\em topological} $n$-simplex: \footnote{In ``fuzzy" simplicial sets, each object $(X[n], p)$ is embedded in an ``expanded" space  $\{(p_0, \ldots, p_n) | \sum_i p_i = -\log p\}$.}
 \[ {\bf \Delta}^n = \{(p_0, \ldots, p_n) \in \mathbb{R}^{n+1} \ | \ 0 \leq p_i \leq 1, \sum_i p_i = 1 \} \]
 The spaces ${\bf \Delta}^n, n \geq 0$ can be viewed as {\em cosimplicial} topological spaces with the following degeneracy and face maps: 
 \[ \delta_i(t_0, \ldots, t_n) = (t_0, \ldots, t_{i-1}, 0, t_i, \ldots, t_n)  \ \mbox{for} \ 0 \leq i \leq n\]
 \[ \sigma_j(t_0, \ldots, t_n) = (t_0, \ldots, t_{j} + t_{j+1}, \ldots, t_n) \ \mbox{for} \ 0 \leq i \leq n\]
 Note that $\delta_i: \mathbb{R}^n \rightarrow \mathbb{R}^{n+1}$, whereas $\sigma_j: \mathbb{R}^n \rightarrow \mathbb{R}^{n-1}$. The equivalence relation $\sim$ above that defines the quotient space is  given as: 
 \[ (d_i(x), (t_0, \ldots, t_n)) \sim (x, \delta_i(t_0, \ldots, t_n) )\]
 \[ (s_j(x), (t_0, \ldots, t_n)) \sim (x, \sigma_j (t_0, \ldots, t_n)) \]
\end{definition}

\section{Summary and Future Work}

To capture categorically the fact that language is replete with superficially different statements, such as ``Charles Darwin wrote" and ``Charles Darwin is the author of", which carry the same meaning, we need to introduce into categorical language models ideas from abstract homotopy theory. We defined an LLM Markov category to represent probability distributions in language generated by an LLM, where the probability of a sentence, such as ``Charles Darwin wrote" is defined by an arrow in a Markov category.  To address this fundamental problem of treating non-isomorphic objects represented by language paraphrases as essentially equivalent, we use categorical homotopy techniques to capture ``weak equivalences" in  an LLM Markov category.  We provided a detailed overview of application of categorical homotopy to LLMs, from higher algebraic K-theory to model categories, building on powerful theoretical results developed over the past half a century. 

We focused primarily on homotopy in this paper, but it is also possible to construct a {\em chain homology} structure on fuzzy simplicial sets that maps an LLM to a sequence of Abelian groups \citep{spivak:fuzzy}. In future work, we want to apply these theoretical tools to  attempts to address language paraphrases in  of empirical LLM studies \citep{khandelwal2020generalizationmemorizationnearestneighbor}, with the goal of developing a deeper understanding of actual deployed LLM systems.

Our paper provides an abstract theoretical analysis of homotopic structure in LLMs. It is not intended as a practical demonstration of how to address the problem of smoothing next-token distributions to reflect semantic similarity between superficially different statements. Such empirical studies have been reported in the literature already, such as $k$-NN LLMs \citep{khandelwal2020generalizationmemorizationnearestneighbor}. The essence of categorical homotopy theory is to capture {\em universal} properties that underlie semantic similarity using the idea of {\em lifting diagrams} \citep{lifting}. It does not address computational issues directly, and it is one of our longer-term goals to bring this type of insight into closer contact with practice.


\end{document}